\documentclass[preprint,12pt]{elsarticle}

\usepackage{tabularx}
\usepackage{url}
\usepackage{booktabs}
\usepackage{ragged2e}
\usepackage{makecell}
\usepackage{multirow}
\newcolumntype{Y}{>{\RaggedRight\arraybackslash}X}

\usepackage{amssymb}
\usepackage{amsmath}
\usepackage{booktabs}

\journal{Robotics and Autonomous Systems}

\begin{document}

\begin{frontmatter}



\title{Factored Reasoning with Inner Speech and Persistent Memory for Evidence-Grounded Human–Robot Interaction}


\author[aff1]{Valerio Belcamino}
\author[aff1]{Mariya Kilina}
\author[aff2]{Alessandro Carfì}
\author[aff1]{Valeria Seidita}
\author[aff2]{Fulvio Mastrogiovanni}
\author[aff1]{Antonio Chella}
\affiliation[aff1]{organization={Department of Engineering, University of Palermo},
            addressline={Viale delle Scienze, 
            \\Bldg. 7}, 
            city={Palermo},
            postcode={90128}, 
            country={Italy}}

\affiliation[aff2]{organization={TheEngineRoom, Department of Informatics Bioengineering, Robotics and System Engineering},
            addressline={via all'Opera Pia 13}, 
            city={Genoa},
            postcode={16145}, 
            country={Italy}}

\begin{abstract}
Dialogue-based human--robot interaction requires robot cognitive assistants to maintain persistent user context, recover from underspecified requests, and ground responses in external evidence, while keeping intermediate decisions verifiable. 
In this paper we introduce \textsf{JANUS}, a cognitive architecture for assistive robots that models interaction as a partially observable Markov decision process and realizes control as a \emph{factored controller} with typed interfaces. 
\footnote{This paper has been submitted to Robotics and Autonomous Systems, and is currently under evaluation.}
To this aim, \textsf{JANUS}
(i) decomposes the overall behavior into specialized modules, related to scope detection, intent recognition, memory, inner speech, query generation, and outer speech, and
(ii) exposes explicit policies for information sufficiency, execution readiness, and tool grounding. 
A dedicated \textit{memory} agent maintains a bounded recent-history buffer, a compact core memory, and an archival store with semantic retrieval, coupled through controlled consolidation and revision policies. Models inspired by the notion of \textit{inner speech} in cognitive theories provide a control-oriented internal \textit{textual flow} that validates parameter completeness and triggers clarification before grounding, while a faithfulness constraint ties robot-to-human claims to an evidence bundle combining working context and retrieved tool outputs. 
We evaluate \textsf{JANUS} through module-level unit tests in a dietary assistance domain grounded on a knowledge graph, reporting high agreement with curated references and practical latency profiles.
These results support factored reasoning as a promising path to scalable, auditable, and evidence-grounded robot assistance over extended interaction horizons.

\end{abstract}



\begin{keyword}


Human-Robot Interaction \sep
Robot Cognitive Architecture \sep
Large Language Model \sep
Multi-Agent Architecture \sep
Inner Speech \sep
Memory
\end{keyword}

\end{frontmatter}



\section{Introduction}
\label{sec:introduction}

The status of interactive, autonomous robots that assist people is translating from controlled demonstrations toward long-term deployment in everyday environments.
They are now being explored and deployed in settings such as healthcare, domestic assistance, or cognitive rehabilitation, and they must support people over days, weeks, or months rather than over a single interaction session~\cite{llm_healthcare, healthcare}. 
This shift from short demos to long-term real-world use raises a broader societal challenge.
To support long-term interaction, robot assistants must not only sustain fluent dialogue but also behave consistently over time, stay aligned with the needs of assisted people, and communicate in ways that promote calibrated trust and appropriate reliance~\cite{llm_trust, healthcare_personality}. 
These requirements are particularly critical in safety-relevant domains in which robots play an advisory, support, or assistance role, such as dietary guidance~\cite{huang2026glenbenchgraphlanguagebasedbenchmark,zhang2025ngqa} and personal care.
In fact, confident but incorrect responses may have direct consequences on the assisted person's well-being.
Furthermore, the quality of the system's explanations can shape humans' reliance on the robot over repeated interactions~\cite{javaid2021explanations, de2025design, trust_health2, trust_health}.

Over the past few years, large language models (LLMs) have emerged as an effective \textit{interface layer} for these systems, because they can translate free-form human-to-robot requests into structured task representations and support multi-step reasoning.
Recent cognitive architectures for robots leverage LLMs as reasoners that produce intermediate steps, and perform simple forms of symbol grounding in their responses via external procedural actions, or \textit{tools}, such as database queries or service-process response requests~\cite{wei2022chain, yao2022react}. 
To deal with uncertainty or ambiguous situations, other approaches explore multiple candidate reasoning paths before committing to an answer, for example, explicitly considering reasoning branches~\cite{yao2023tree}.
In a more system-oriented perspective, the possibility of connecting LLMs to broader tool ecosystems and multi-modal components has made it possible to design robot architectures for assistants capable of routing sub-tasks to specialized resources and generating executable policies for the specific embodiment~\cite{liu2023g, shen2023hugginggpt, liang2022code}.

Despite such relevant positive results, long-term interaction imposes constraints that are not addressed by \textit{prompt engineering} alone.
A robot assistant that interacts reliably over time must 
(i) detect when essential information is missing, 
(ii) distinguish retrieved evidence from unsupported inferences, and 
(iii) decide when to ask the human for clarification rather than proceed with an under-specified plan~\cite{shinn2023reflexion, madaan2023self, wataoka2024self}. 
It should be pointed out that reliability not only depends on final answers, but on intermediate steps as well, since wrong intent selection, incorrect tool choice, irrelevant retrieval, or mismatched explanations can each jeopardize the interaction even if the output is fluent and realistic~\cite{zheng2023judging, kocmi2023large}. 
In long-term interactions, intermediate outputs can also have cumulative effects.
They may degrade a stored user model, reinforce incorrect assumptions, and ultimately distort trust through over-reliance on confident but ungrounded content.
Long-term interaction is inherently multi-domain.
Tasks, constraints and context shift over time, and reasoning performance can degrade under distribution shifts inherent to different interaction domains~\cite{belcamino2026generalizationgapllmplanning}. 
These considerations motivate explicit architectural design choices, rather than relying on scaling the model capacity alone.

In this paper, we propose \textsf{JANUS}, a cognitive architecture for assistive robots centered on two central constructs, namely \emph{inner speech} and \emph{memory}.
Cognitive accounts describe ``inner speech'' as an internalized form of language used to regulate behavior, monitor decisions and support reasoning, closely linked to working-memory mechanisms and self-monitoring~\cite{alderson2015inner, fernyhough2023inner}.
In engineering terms, inner speech can be implemented as an explicit \emph{control channel} that structures deliberation for verification and decision-making.
In \textsf{JANUS}, inner speech is an internal \textit{textual flow} used to validate intent, check reasoning constraints, and decide whether the robot has enough information to act reliably.
Critically, this control channel need not be directly exposed to the assisted person.
Instead, it can serve to govern downstream decisions about tool use and the interaction flow.
In parallel, memory research emphasizes that persistence is not only storage but above all organization and revision of beliefs~\cite{squire1995retrograde, nader2000labile, tayler2013new}.
For the design of robot assistants, such insight implies that memory could be treated as a managed component.
In \textsf{JANUS}, memory persistence is used to select what to consolidate at different timescales, resolve conflicts, and apply explicit update policies rather than accumulating unstructured history.

This paper presents the main facets of \textsf{JANUS}, and focuses on its cognitive constructs related to inner speech and memory, while describing its factored reasoning architecture designed for long-term, multi-domain interaction.
This work stems by a multi-level expansion of previous work of ours, which builds on an initial prototype for dietary guidance~\cite{belcamino2025social}. 

\textsf{JANUS} uses \emph{inner speech} as an assessment stage to validate the intents of the person it interacts with, check the completeness of the provided information, and decide when to \textit{actively} seek additional information.
It leverages \emph{outer speech} as a separate channel to acquire information from the assisted person when needed, for example in the form of answers, explanations, and clarifications to questions grounded in the retrieved evidence. 
The separation between inner and outer speech serves the purpose of improving the robustness of long-term interaction, since internal deliberation can be optimized for verification and the decision-making process, whereas the external communication can be optimized for clarity and the calibration of trust. 

From a software engineering perspective, \textsf{JANUS} supports online domain switching through a dedicated \textit{customization} layer.
Such layer encapsulates domain schemas, tools, prompts, and database connectors.
This structure enables \textsf{JANUS} to operate across different domains with the same core reasoning logic. 
Moreover, it provides scalable persistence via a specialized memory \textit{agent} endowed with hierarchical consolidation, which organizes information across multiple timescales, and applies controlled updates to maintain coherence over time. 
Such a \textit{role separation} follows the nowadays broader view that reliable robot assistants may benefit from explicit role decomposition across reasoning, tool use, verification, and explanation, rather than relying on a single undifferentiated generation step~\cite{besta2025reasoning}.

The main contributions of this paper are:
(i) a modular, memory-augmented factored reasoning robot cognitive architecture for long-term interaction, which explicitly separates \emph{inner speech} (control) from \emph{outer speech} (communication);
(ii) a dialogue management interface that enables run-time multi-domain switching by encapsulating domain-specific schemes, tools, and connectors while keeping core reasoning modules unaffected;
(iii) a dedicated memory agent with hierarchical consolidation capabilities, which treats persistence as a managed, bounded module with explicit update policies.

The remainder of the paper is structured as follows. 
Section~\ref{sec:state_of_the_art} reviews related work on multi-agent architectures, in so far as factored reasoning in our case is modeled as a \textit{weak} multi-agent framework, and methods to model memory-like capabilities for robot cognition. 
Section~\ref{sec:proposed_architecture} introduces first a conceptual model of \textsf{JANUS}, then describes it system architecture, which emphases the role of inner speech, and finally focuses on its  memory capabilities as well as its consolidation/update mechanisms. 
Section~\ref{sec:experimental_evaluation} reports implementation notes, our use case, validation results, and a discussion. 
Conclusions follow.

\section{State of the Art}
\label{sec:state_of_the_art}

This Section reviews prior work that motivates our architecture along two axes.
On the one hand, we review LLM-driven agent-based pipelines for deliberation and tool-grounded action in robotics, which have inspired the design of our factored reasoning pipeline.
On the other hand, we focus on cognitive and engineered approaches to model relevant memory traits for long-term human-robot interaction.
Across both axes, a recurring methodological tension emerges between \emph{expressiveness}, that is, powerful language-driven reasoning and tool use, and \emph{control}, that is, bounded state, verifiable intermediate decisions, and faithful communication with the assisted person.

\paragraph{Multi-agent architectures}
Despite decade-old advances in the field of robot cognitive architectures, robots \textit{brains} are commonly organized as modular pipelines, for example, adhering to the well-established Sense--Plan--Act paradigm, to support extensibility and fault isolation.
Recent cognitive architectures based on LLMs extend these modularity traits to language-mediated planning, in which an LLM-based model decomposes a human request into sub-steps, selects actions, and produces structured outputs that can be executed via a set of downstream procedures.
It has been emphasized that encouraging an LLM-based model to make intermediate reasoning steps explicit in the output, rather than yielding only the final answer, could improve controllability and problem solving~\cite{wei2022chain}.
A similar line of work strives to couple reasoning and action.
Approaches in the literature aim at delivering models capable to alternate between short internal deliberation and explicit \textit{tool} invocations, whereas ``tools'' are extra-model procedures, such as database queries or Application Programming Interface (API) requests, so that the final model output can be grounded, at least partially, in externally retrieved results~\cite{yao2022react}.
When a human request admits multiple plausible interpretations, search-based reasoning frameworks have been leveraged to explore alternative, intermediate trajectories before committing to a final decision, which may help recover from early misunderstandings~\cite{yao2023tree}.
While these approaches improve tool grounding and reduce some failure modes, they typically leave implicit \emph{which} intermediate decisions must be verified, \emph{when} the system should ask for clarification, and \emph{how} verification outcomes constrain the human-robot communication process.

With a specific focus related to tool use, approaches in the literature train models to decide when to invoke tools and how to format tool inputs as part of the generation process~\cite{schick2023toolformer}.
This idea can be scaled up by connecting a model to large tool catalogs, and translating natural-language goals into structured tool calls~\cite{liu2023g}.
Orchestration frameworks generalize this pattern by treating a model as a \textit{controller} that routes sub-tasks to specialized tools or models, which may be useful when robot assistants are expected to combine perception, retrieval, planning, and dialogue management~\cite{shen2023hugginggpt}.
As examples of frameworks explicitly used in robotics, code-synthesis approaches have been used to generate executable policies from language descriptions. This demonstrates a direct interface between high-level intent and robot control logic~\cite{liang2022code}.

From the perspective of long-term autonomy, these approaches highlight the value of role decomposition.
However, they also raise a system-level question, that is, where is the \emph{control state} that decides readiness to act, enforces constraints, and governs when additional information must be requested?

In fact, interaction over long temporal horizons requires mechanisms to detect and correct failures.
Self-reflection and self-refinement strategies leveraged prior errors or critiques to improve subsequent outputs~\cite{shinn2023reflexion, madaan2023self}, whereas self-optimization based on rewards or preference has shown improvement in the average performance~\cite{wataoka2024self}.
More recently, the evaluation of \textit{agentic} LLM-based pipelines has been integrated into system design workflows. Intermediate artifacts, such as agent traces (for example, intent recognition decisions, tool invocations, retrieved database results, and explanations provided to assisted users), can be automatically assessed using rubric-based scalar ratings, including LLM-as-a-judge evaluation protocols~\cite{zheng2023judging, kocmi2023large}.
Recent results show that planning behaviors can degrade under domain or constraint shifts, which reinforces the need for explicit architectural controls rather than the sole reliance on emergent generalization~\cite{belcamino2026generalizationgapllmplanning}.

These trends suggest that the explicit availability of \emph{intermediate decisions} serves a critical role: they constitute the \textit{locus} at which robustness, safety, and long-term reliability can be systematically enforced, provided that they are explicitly represented and regulated by a principled control loop.

In many agent-based architectures, internal deliberation and external explanations are produced by the \textit{same} generation step. 
At first glance, this may appear beneficial, as a single generation step can yield explanations that are \textit{linguistically consistent} with the model’s internal reasoning. 
However, this coupling provides limited guarantees of \emph{faithfulness} to the model’s actual behavior.
In particular, when deliberation, tool selection, and explanation are co-generated, explanations may become fluent \emph{post-hoc} rationalizations that are not explicitly constrained by what the model actually retrieved, verified, or enforced. 
As a result, an explanation may remain coherent with the generated reasoning trace while still diverging from execution traces -- such as tool outputs, retrieved evidence, or constraint checks -- making it difficult to guarantee traceability and to support calibrated trust~\cite{javaid2021explanations,de2025design,10731451,10.1145/3656650.3656675}. 
This issue is especially critical in human–robot interaction, where explanation quality directly affects trust calibration~\cite{verbal_trust, behavior}.

This motivates the design of robot cognitive architectures that explicitly separate internal control decisions (for example, interpretation, verification, and execution readiness) from the external communication channel used to generate human-facing explanations, thereby distinguishing \emph{what the system uses to decide} from \emph{what the system chooses to say}.

\paragraph{Approaches to model memory}
Cognitive accounts of memory emphasize a number of aspects related to the intrinsic behavior of memory \textit{systems} that are of the utmost importance to synthesize an operational view of memory.
Such memory systems are constrained by limited time, attention, and storage, and are governed by multiple control processes. 
Theories about working memory distinguish limited-capacity stores from longer-term representations, and highlight internal rehearsal mechanisms supporting maintenance and manipulation over short time horizons~\cite{alderson2015inner, fernyhough2023inner}. 
Research on long-term memory further indicates that persistence requires stabilization over time.
Accounts of consolidation processes describe how new traces become durable and are retained reliably over time~\cite{squire1995retrograde,tayler2013new}, whereas findings on reconsolidation show that retrieved content can become temporarily modifiable and open to purposeful revision. This suggests that updates must be handled carefully to avoid incoherence when new information revises older content~\cite{nader2000labile}.
Together, these perspectives motivate engineering policies that 
(i) keep short-term interaction context bounded, 
(ii) promote stable information such as user preferences into compact representations, and 
(iii) revise stored content when humans provide corrections.

A design-level implication for robot assistants is that memory should not be treated as a passive \textit{log}, but as a controlled resource whose \emph{writes} and \emph{updates} must be deliberate and selective.

Inner speech theories link the internal dialogue we all are aware of to internal process monitoring and executive control.
This link seems to suggest the case of design choices for robot cognitive architectures in which an internal verbal state operates as a control signal rather than as a mere source of outward explanations~\cite{alderson2015inner, fernyhough2023inner}. 
Approaches in the literature seek to operationalize these ideas.
For example, the approaches described in~\cite{chella2020developing} maintain an internal verbal state that persists across dialogue turns and sessions, and is updated over time to coordinate perception, memory retrieval, and action selection. 
Related work suggests that recurrent self-dialogue can support learning and coordination in artificial agents~\cite{pipitone2021robots}. 
It has been shown that selectively externalizing internal reasoning can contribute to shape in humans the perceptions of the robot to a great extent, for example for what concerns intellectual capabilities and \textit{animacy}, which motivates the availability of principled control mechanisms over both content and timing of disclosures~\cite{pipitone2024robot, relatable, relatable2, transparency, embodiment}.

This line of research points to a key design lever for long-term interaction.
Internal linguistic representations can be used to stabilize and regulate behavior without necessarily being exposed \textit{verbatim} to assisted humans.

In LLM-based architectures, retrieval-augmented generation (RAG) grounds outputs by conditioning on retrieved documents rather than relying solely on parametric memory, that is, information stored in model's weights~\cite{lewis2020retrieval}.
In case of multi-hop questions and heterogeneous sources of information, graph-based retrieval and structured representations like knowledge graphs and schemas aim to improve coherence and controllability by making relations semantically explicit~\cite{besta2024graph, zeng2024structural, peng2025graph}. 
Agent-centric memory streams add summarization and reflection to compress long interaction histories into compact states, which can be used for reasoning and tool selection~\cite{park2023generative}. 

In robot cognitive architectures, these approaches align with engineering constraints.
Memory should remain bounded, prioritize relevance and recency, and support revision when humans update preferences or constraints.

These strands motivate two design principles for robot cognitive architecture aimed at long-term human-robot interaction: 
(i) internal control and verification should be separated from outward communication to preserve faithfulness and enforce the calibration of trust, and 
(ii) memory should be implemented as a managed module with explicit consolidation and update policies, rather than as unbounded prompt history.
These themes motivate approaches that treat long-horizon interaction as a coupled problem of \emph{control} (deliberation, verification, readiness) and \emph{persistence} (bounded state, consolidation, revision), rather than as a single-step generation process.

\section{\textsf{JANUS}: a Factored Reasoning Architecture for Long-Term Interaction}
\label{sec:proposed_architecture}

\begin{figure}[t!]
\centering
\includegraphics[width=\linewidth]{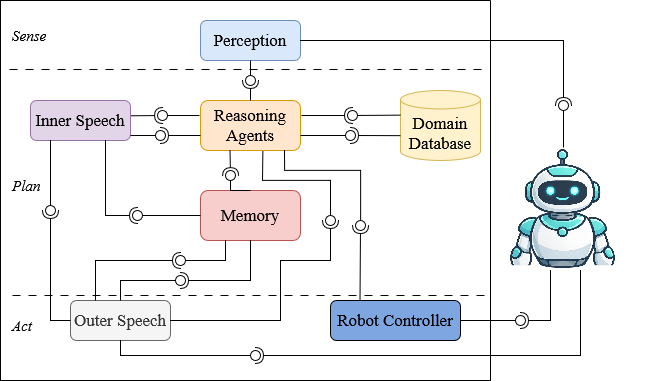}
\caption{
Overview of \textsf{JANUS} as a classical Sense-Plan-Act pipeline.
The \textit{Sense} layer acquires and processes input from the human via a series of perception modules.
The \textit{Plan} layer integrates reasoning modules, inner speech, domain databases for retrieval and query grounding, as well as a dedicated memory agent that maintains a persistent interaction context.
The \textit{Act} layer externalizes decisions through outer speech for communications with the human, and through the robot controller for action execution.}
\label{fig:architecture}
\end{figure}

\subsection{Conceptual Model}
\label{sec:agent_model}

This Section introduces an abstract conceptual model which aims at formalizing long-horizon interaction and motivates the underlying design principles of \textsf{JANUS}.
We describe the robot cognitive architecture as a \emph{factored controller} composed of logical modules (which can be mapped to one or more ``agents'' for convenience) with typed interfaces, and shown in Figure~\ref{fig:architecture}.
These modules are not assumed to be independent decision-makers.
Instead, they represent a structured decomposition of the overall policy governing \textsf{JANUS} into interpretable intermediate variables, for example, domain selection, intent schemas, readiness signals, evidence bundles, and memory updates.
The concrete system instantiation and module implementations are described in Section~\ref{sec:system_architecture}, whereas the memory module and its update/consolidation mechanisms are detailed in Section~\ref{sec:memory_model}.

In \textsf{JANUS}, we model long-term human-robot interaction as a partially observable Markov decision process (POMDP)
\[
\mathcal{M}=\langle \mathcal{Z},\mathcal{A},\mathcal{O},T,Z,R,\gamma\rangle,
\]
which provides a formal scaffold for long-horizon robot assistant behavior under uncertainty.
The interaction process unfolds over a series of turn-taking steps, which for the sake of modeling are considered to be collapsed in a series of time instants $t$.
At each turn, the human provides an input request $x_t\in\mathcal{X}$, which is captured by the \textit{Perception} module, for example, an utterance which yields an observation $o_t\in\mathcal{O}$ derived from $x_t$ and the available context.
Then, the robot assistant executes an action $a_t\in\mathcal{A}$ that may include clarification, tool invocation, or memory operations, by means of an internal confabulation managed by the \textit{Inner Speech} agent.
The robot assistant produces an output $y_t\in\mathcal{Y}$ via one or more \textit{Reasoning} agents, for example, an answer, an explanation, or a clarification question via the \textit{Outer Speech} agent, optionally coupled with an action command for the robot to execute via the \textit{Robot Controller} module.

It should be emphasized that we introduce this POMDP formulation to motivate the architecture and its control variables.
Here, $T$ denotes the (possibly unknown) transition dynamics over the latent interaction state, which captures in principle both the evolution of the human/environment state, and the controlled update of the robot assistant's internal state (including memory updates as formalized later), whereas $Z$ denotes the observation model that captures partial observability of goals, constraints, and preferences from language and context.
We include $R$ and $\gamma$ for completeness of the common POMDP formulation, but our focus in this paper is architectural, that is, we do not assume that \textsf{JANUS} solves the POMDP optimally, nor do we require an explicit specification of a reward function.

Let $z_t \in \mathcal{Z}$ denote the latent interaction state, that is, the POMDP state, at turn $t$, which summarizes control-relevant variables only partially observable given the observations derived from $x_t$ and the available context.
In \textsf{JANUS}, $z_t$ includes at least:
(i) the active domain $d_t \in \mathcal{D}$,
(ii) the underlying goal $g_t \in \mathcal{G}(d_t)$ of the human who is interacting with the robot,
(iii) a structured intent schema $\tau_t \in \mathcal{T}(d_t)$,
(iv) a (possibly partial) parameter assignment $\theta_t$ for $\tau_t$, and 
(v) a persistent \textit{user context} (preferences, constraints, identifiers) accumulated over long time horizons.
We model such persistent context through an explicit memory state $S_t$, which is queried and updated across turns and is therefore a key component of $z_t$ in long-horizon interaction (detailed in Section~\ref{sec:memory_model}), such that
\begin{equation}
S_t=\langle H_t, C_t, A_t\rangle,
\end{equation}
where 
$H_t$ is the bounded historic interaction buffer, 
$C_t$ is a compact core memory, and 
$A_t$ is an archival memory store.

For each turn, \textsf{JANUS} constructs a bounded working context $W_t$ from memory.
As it will be described more thoroughly in Section~\ref{sec:memory_model}, a key decision is whether the current request can be addressed using the already available context (core memory and recent history), or whether archival retrieval is needed.
We make this decision by recurring to an information-sufficiency gate $s_t$, modeled as a binary predicate, such that
\begin{equation}
s_t = \mathrm{SUFF}(x_t, C_t, H_t) \in \{0,1\},
\label{eq:suffgate}
\end{equation}
where $\mathrm{SUFF}$ is implemented as a classifier (which, in \textsf{JANUS}, is LLM-based) estimating whether $x_t$ can be handled using $(C_t, H_t)$ alone.
If $\mathrm{SUFF}$ returns $1$, the \textit{Memory} agent synthesizes $W_t$ from $(C_t, H_t)$.
Otherwise, if it returns $0$, the \textit{Memory} agent augments $W_t$ with a Top-$k$ retrieval from the archival memory $A_t$.
This explicit gate makes memory access \emph{controlled} and \emph{auditable} rather than an unbounded prompt-history accumulation.

Each domain $d \in \mathcal{D}$ provides a library of task schemas $\mathcal{T}(d)$, each with a typed parameter space.
We represent \textit{intent} as a pair $(\tau_t, \theta_t)$ with $\tau_t\in\mathcal{T}(d_t)$ and $\theta_t$ a typed (possibly partial) assignment.
We define a \emph{completeness} indicator, modeled as a predicate, such that
\begin{equation}
p_t = \mathrm{COMP}(\tau_t, \theta_t)\in\{0,1\},
\label{eq:comp}
\end{equation}
which returns $1$ \textit{iff} all required parameters for $\tau_t$ are present \textit{and} satisfy basic type/constraint checks.
This predicate formalizes the notion of execution readiness that \textit{Inner Speech} assesses before allowing for \textit{grounded} action.

When the task requires any grounding external to the model, for example in the form of database queries, \textsf{JANUS} produces a query specification $q_t$ and obtains tool evidence $E_t$ by executing $q_t$ through specific domain adapters, as it will be described in Section \ref{sec:system_architecture}.
In this way, \textsf{JANUS} explicitly separates \emph{memory sufficiency}, that is, whether archival retrieval is needed at all, from the need for \emph{tool grounding}.
To this aim, we model a binary tool-grounding indicator $\rho_t$, and we formalize it as a predicate:
\begin{equation}
\rho_t = \mathrm{TOOL}(d_t,\tau_t,\theta_t,W_t)\in\{0,1\},
\label{eq:toolneed}
\end{equation}
where $d_t$ is the active domain, $(\tau_t,\theta_t)$ is the structured task representation, and $W_t$ is the bounded working context returned by the memory subsystem.
Intuitively, $\mathrm{TOOL}$ returns $1$ when generating a faithful response requires external evidence, while it returns $0$ when the response can be generated from the bounded working context alone.
We observe that each task schema induces an evidence requirement set $\mathcal{R}(\tau_t,\theta_t)$, for example, for required records or attributes), and therefore we define
\begin{equation}
\mathrm{TOOL}(d_t,\tau_t,\theta_t,W_t) \;=\; \neg \mathrm{SAT}\!\left(W_t,\mathcal{R}(\tau_t,\theta_t)\right),
\end{equation}
that is, external tools are invoked when the working context does not already satisfies the evidence requirements of the current task, conveniently modeled as a satisfiability predicate $\mathrm{SAT}$.
When $\rho_t=0$, we set $q_t=\emptyset$ and $E_t=\emptyset$.

From the qualitative description of such an abstract model, therefore, it emerges that rather than modeling \textsf{JANUS} as a monolithic mapping $\pi(y_t \mid x_{1:t})$, we define a factored controller composed of specialized agent-like modules with typed interfaces.
Equations~\eqref{eq:scope}--\eqref{eq:outerspeech} specify a \emph{factorization} of the overall input-output mapping into typed intermediate variables and their dependencies, forming a \emph{computational graph} that represents the control loop collectively implemented by the modules and agents in \textsf{JANUS}.
While it should be emphasized that the sequence below is \emph{not} an algorithmic \textit{pseudocode}, it is noteworthy that the ordering reflects dependency constraints, that is, which variables require which inputs, while the concrete execution could be implemented sequentially, asynchronously, or with caching, as it will be described in Section~\ref{sec:system_architecture}.
At each turn, we define the following intermediate variables:
\begin{align}
\hat d_t &= f_{\mathrm{SD}}(x_t, d_{t-1}), \label{eq:scope}\\
(\tau_t,\tilde\theta_t) &= f_{\mathrm{IR}}(x_t,\hat d_t), \label{eq:intent}\\
\theta_t &= f_{\mathrm{PP}}(\tau_t,\tilde\theta_t,\hat d_t), \label{eq:postproc}\\
(s_t, W_t) &= f_{\mathrm{Mem}}(x_t, S_t), \label{eq:mem}\\
(c_t,\rho_t) &= f_{\mathrm{IS}}(x_t,\hat d_t,\tau_t,\theta_t,W_t), \label{eq:innerspeech}\\
q_t &= f_{\mathrm{QG}}(x_t,\hat d_t,\tau_t,\theta_t,W_t,c_t)\ \ \text{if }\rho_t=1, \label{eq:qgen}\\
E_t &= \mathrm{Exec}(\hat d_t, q_t)\ \ \text{if }\rho_t=1, \label{eq:exec}\\
y_t &= f_{\mathrm{OS}}(x_t,\hat d_t,\tau_t,\theta_t,W_t,c_t,\rho_t,E_t), \label{eq:outerspeech}
\end{align}
where
$f_{\mathrm{SD}}$ refers to a \textit{Scope Detection} agent, 
$f_{\mathrm{IR}}$ to an \textit{Intent Recognition} agent,
$f_{\mathrm{PP}}$ to an \textit{Intent Post-processing} agent, 
and $f_{\mathrm{Mem}}$ implements a \textit{Memory} agent using the gate in Equation \eqref{eq:suffgate}.
Moreover, 
$f_{\mathrm{IS}}$ denotes the \textit{Inner Speech} agent, 
$f_{\mathrm{QG}}$ the \textit{Query Generation} agent, 
$\mathrm{Exec}(\cdot)$ executes tool calls through domain backends, and 
$f_{\mathrm{OS}}$ is the \textit{Outer Speech} agent, for robot-to-human communication.

\textit{Inner Speech} returns a control variable $c_t$ that governs the interaction flow.
We model it as a finite control alphabet, such that
\begin{equation}
c_t \in \mathcal{C}=\{\textsc{Proceed},\ \textsc{Clarify},\ \textsc{SwitchDomain},\ \textsc{Reject}\},
\label{eq:control}
\end{equation}
optionally augmented with a set of missing slots $M_t$ and validation flags.
Operationally, as anticipated in Equation~\eqref{eq:comp}, \textit{Inner Speech} enforces an execution-readiness condition, that is,
\begin{equation}
c_t = \textsc{Proceed} \ \Rightarrow\ \mathrm{COMP}(\tau_t, \theta_t) = 1,
\label{eq:readiness}
\end{equation}
while $c_t = \textsc{Clarify}$ indicates that the request is under-specified and \textsf{JANUS} should ask a targeted clarification question rather than proceed with grounding.
This explicit gating prevents the robot assistant from acting in under-specified conditions and from accumulating unverified assumptions into memory.

As we mentioned before, a key principle in \textsf{JANUS} is that internal control decisions and communication acts with the assisted person are the result of the integrated behavior of interacting agents.
The signal $c_t$ of Equation~\eqref{eq:control} can condition the output $y_t$, but need not be externalized \textit{verbatim}.
This design choice enables the optimization of internal deliberation for verification and readiness, while optimizing outward communication for clarity and the calibration of trust in the assisted human.
Formally, we treat such design choice as a separation between a \emph{decision policy} (related to upstream agents and \textit{Inner Speech}) and a \emph{communication policy} (related to \textit{Outer Speech}).

Let $B_t = (W_t, E_t)$ denote the \emph{evidence bundle} available at turn $t$, that is, the bounded working context from memory and (when applicable) tool evidence.
Let $\mathrm{Claims}(y_t)$ be the set of atomic claims expressed in $y_t$.
Let $\mathrm{Supp}(B_t)$ denote the set of claims supported by $B_t$ under domain semantics.
For turns that proceed with an answer to the human, such that $c_t = \textsc{Proceed}$, we target the constraint:
\begin{equation}
\mathrm{Claims}(y_t)\subseteq \mathrm{Supp}(B_t)\ \cup\ \mathrm{SafeDefaults},
\label{eq:faithfulness}
\end{equation}
where $\mathrm{SafeDefaults}$ contains explicitly marked statements that do not require grounding, for example, uncertainty disclosures, generic safe advice, or clarification requests.
Equation~\eqref{eq:faithfulness} formalizes the requirement that what \textsf{JANUS} communicates should be anchored to what it actually retrieved or verified, rather than to unconstrained post-hoc rationalization.

At the end of each turn, memory is updated through a managed operator:
\begin{equation}
S_{t+1}=\mathcal{U}(S_t, x_t, y_t, \hat d_t, \tau_t, \theta_t, c_t, s_t, \rho_t, E_t),
\label{eq:memupdate}
\end{equation}
where $\mathcal{U}$ implements bounded historic insertion and capacity-triggered consolidation/revision, as it will be described in Section \ref{sec:memory_model}.
This makes \textit{memory persistence} an explicit component governed by update policies, rather than an unbounded accumulation of interaction traces.


As we posited while discussing relevant approaches in the literature, in human-robot interaction processes spanning long temporal horizons, often the decisive factor is the balance between intermediate, \emph{internal} decisions preceding the observable answer, that is, domain assignment, schema selection, slot completeness, evidence acquisition, and the choice to \textit{ask} rather than to \textit{assume}.
This motivates the choices done in \textsf{JANUS}, which makes \emph{sufficiency} ($s_t$), \emph{readiness} ($c_t$), and \emph{persistence} ($\mathcal{U}$) relevant variables of the control loop.

\begin{figure}[t!]
\centering
\includegraphics[width=\linewidth]{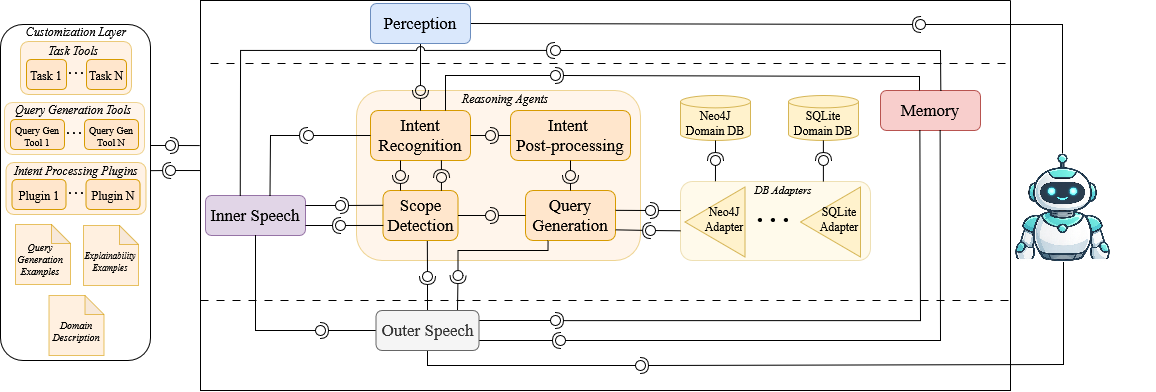}
\caption{
Overview of \textsf{JANUS} as a factored reasoning and control architecture.
A \textit{Perception} module forwards the input collected by the human to a set of \textit{Reasoning} modules.
Among them, 
\textit{Intent Recognition} and \textit{Intent Post-processing} generate a structured task representation from the user intent, whereas
\textit{Scope Detection} estimates the active interaction domain.
If needed, \textit{Query Generation} and database adapters acquire external information. 
This process is supervised by \textit{Inner Speech}, which performs execution-readiness checks.
The \textit{Memory} agent provides persistent context and a bounded working context for the current turn-taking step, while 
\textit{Outer Speech} generates responses for the human, and may interface with the robot controller for the execution of embodied actions.
A runtime \textit{Customization Layer} module binds domain-specific task schemas, tools, plugins, and prompt assets without modifying the core pipeline.
}
\label{fig:implementation}
\end{figure}

\subsection{System Architecture and Information Flow}
\label{sec:system_architecture}

This Section describes the system-level realization of the factored controller introduced in Section~\ref{sec:agent_model}.
As we pointed out above, \textsf{JANUS} is implemented as a memory-augmented distributed reasoning pipeline that transforms perceptual input into grounded, evidence-backed responses through the control loop formalized in Equations~\eqref{eq:scope}--\eqref{eq:outerspeech}.
Therein, we introduced such mappings as $f_{\mathrm{SD}}, f_{\mathrm{IR}}, f_{\mathrm{IS}}, f_{\mathrm{QG}}, f_{\mathrm{OS}}$, which denote \emph{functional roles} in the factored controller.
The related modules can be seen as agents in a weak sense, employing specialized decision rules while interacting in ways that support the governance exerted by \textit{Inner Speech}. 
In contrast, while \textit{Inner Speech} acts as a control module, \textit{Memory} functions as a true agent, that is, a stateful component governed by explicit update policies.

The current \textsf{JANUS} pipeline instantiates the functional roles in Equations~\eqref{eq:scope}--\eqref{eq:outerspeech} as agents encapsulated within software modules, that is, \textit{Scope Detection}, \textit{Intent Recognition}, \textit{Intent Post-processing}, \textit{Query Generation}, \textit{Inner Speech}, and \textit{Outer Speech}, together with a persistent \textit{Memory} agent and an adapter layer for grounded execution, see Figure~\ref{fig:implementation}.

The interaction begins when a request originated by the human is captured by the \textit{Perception} module.
\textit{Scope Detection} determines whether the request belongs to the currently active conversation domain, or whether a domain switch is required, therefore yielding the domain estimate $\hat d_t$ of Equation~\eqref{eq:scope}.
Given the active domain, \textit{Intent Recognition} maps the request to a structured task representation $\tau_t$, and extracts a typed (possibly partial) parameter set $\tilde\theta_t$, as in Equation~\eqref{eq:intent}.
\textit{Intent Post-processing} then refines this representation into $\theta_t$ to meet implementation requirements, such as domain-specific normalization, constraint checks, and validation, as foreseen in Equation~\eqref{eq:postproc}.

The \textit{Memory} agent provides persistent context across turns and domains through an explicit memory state $S_t =\langle H_t, C_t, A_t \rangle$, as it will be detailed in Section~\ref{sec:memory_model}.
At each turn, this agent constructs a bounded working context $W_t$, and exposes the information-sufficiency gate $s_t$ of Equation~\eqref{eq:suffgate}, which controls whether archival retrieval from $A_t$ is required in addition to $(C_t, H_t)$.

As seen in Equation~\eqref{eq:innerspeech}, \textit{Inner Speech} acts as an explicit internal control stage.
It validates the current interpretation, checks parameter completeness according to Equation~\eqref{eq:comp}, and produces a control decision $c_t$ that governs the interaction flow, as detailed in Equation~\eqref{eq:control}, for example, whether \textsf{JANUS} should proceed or ask for clarification.
In addition, the agent outputs the tool-grounding indicator $\rho_t$ of Equation~\eqref{eq:toolneed}, which determines whether external evidence acquisition is required.
If the request remains underspecified, \textit{Inner Speech} triggers a clarification turn via \textit{Outer Speech} rather than proceeding with incomplete grounding.

If $c_t = \textsc{Proceed}$ and $\rho_t = 1$, then \textit{Query Generation} produces one or more executable query specifications $q_t$ aligned with the selected task schema and its typed arguments, as shown in  Equation~\eqref{eq:qgen}.
Queries are executed through domain adapters to obtain tool evidence $E_t$, see Equation~\eqref{eq:exec}.
Then, \textit{Outer Speech} generates a response for the human as well as an explanation conditioned on the evidence bundle $B_t = (W_t, E_t)$, as in Equation~\eqref{eq:outerspeech}, and can interface with the robot controller when applicable.
At the end of each turn, the \textit{Memory} agent updates $S_t$ according to its consolidation and revision policy shown in Equation~\eqref{eq:memupdate}, which preserves continuity across turns and domain switches.

A central design principle of \textsf{JANUS} is that conversation domains are treated as explicit runtime states.
Domain-specific behavior is configured at runtime through a \textit{Customization Layer} rather than hard-coded into the reasoning modules.
Each domain is specified by
(i) a \textit{Domain Description}, 
(ii) a set of \textit{Task Tools} with typed parameter schemas, and
(iii) a series of supporting assets constraining how tasks are interpreted, grounded, and explained.
In our implementation, Task Tools correspond to structured task definitions that encode both the task description itself and the typed argument space.
Beyond task schemas, the \textit{Customization Layer} bundles domain-specific resources required for grounding and communication.
For example, it binds \textit{Query Generation} tools that map structured tasks to query templates or multi-step retrieval strategies, as well as plugins implementing domain-specific post-processing behavior.
Furthermore, it binds prompt assets, for example, \textit{Query Generation} and \textit{Outer Speech} examples, which standardize formatting and encourage evidence-grounded explanations.

When \textit{Scope Detection} triggers a domain switch, the active set of Task Tools, \textit{Query Generation} tools, plugins, and prompt assets is replaced with the configuration of the selected domain.
The \textit{Memory} agent remains shared across domains.
Therefore, persistent user context such as identifiers, preferences, and constraints remains available even when the dialogue with humans interleaves multiple topics.

\begin{figure}[t]
\centering
\includegraphics[width=\linewidth]{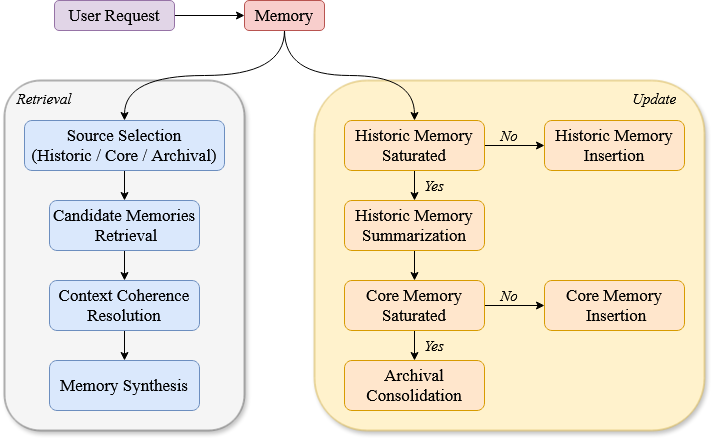}
\caption{
A view of the \textit{Memory} agent as a controlled retrieval-update subsystem.
At each turn, the agent (on the left hand side) constructs a bounded working context $W_t$ from the memory state $S_t = \langle H_t, C_t, A_t \rangle$ via the information-sufficiency gate $s_t$ of Equation~\eqref{eq:suffgate}, and the optional archival retrieval.
The resulting $W_t$ is consumed by upstream modules, for example, \textit{Inner Speech} and \textit{Query Generation}, and contributes to the evidence bundle $B_t = (W_t, E_t)$.
At the end of the turn (on the right hand side), the agent applies a managed update operator $\mathcal{U}$, as described in Equation~\eqref{eq:memupdate}, inserting new traces into $H_t$ and triggering capacity-based consolidation and revision into $C_t$ and $A_t$.
}
\label{fig:memory_agent}
\end{figure}

\subsection{The Memory Agent}
\label{sec:memory_model}

Long-horizon interaction in \textsf{JANUS} is supported through a dedicated \textit{Memory} agent that maintains a shared persistent memory state and exposes two typed interfaces:
(i) a \emph{retrieval} interface returning a bounded working context for the current turn, and 
(ii) an \emph{update} interface applying controlled persistence at the end of the turn.
This Section formalizes the memory state, the working-context construction process, and the consolidation/revision mechanisms used to keep persistence bounded and coherent.

Consistently with Section~\ref{sec:agent_model}, we formalize the memory state at a given interaction turn $t$ as $S_t = \langle H_t, C_t, A_t \rangle$, where 
$H_t$ is a bounded historic buffer of recent interaction traces, 
$C_t$ is a compact set of consolidated core \textit{facts}, for example stable preferences, identifiers, constraints, and 
$A_t$ is an archival store supporting long-term retrieval.
In \textsf{JANUS}, memory is a control-relevant resource.
The working context returned by the \textit{Memory} agent can be used to fill missing task parameters, and the controlled update operator of Equation~\eqref{eq:memupdate} prevents unverified assumptions from being accumulated as persistent facts.

At each turn, the \textit{Memory} agent constructs a bounded working context $W_t$ that summarizes the information made available to the controller for decision making.
As described in Section~\ref{sec:agent_model}, the first decision is whether the current request can be handled using core memory and recent history only, or whether archival retrieval is required.
We model this decision through the information-sufficiency gate $s_t = \mathrm{SUFF}(x_t, C_t, H_t) \in \{0,1\}$, where the predicate $\mathrm{SUFF}$ acts as a binary classifier estimating whether $x_t$ can be addressed using $(C_t,H_t)$ alone.
If $s_t = 1$, the \textit{Memory} agent synthesizes \textit{locally} a working context from recent history and core facts, such that
\begin{equation}
W_t = \Phi_{\mathrm{local}}(x_t, C_t, H_t).
\end{equation}
Otherwise, the agent additionally retrieves a bounded set of archival items and \textit{augments} the working context accordingly, that is
\begin{equation}
W_t = \Phi_{\mathrm{aug}}\!\left(x_t, C_t, H_t, \mathrm{RETRIEVE}_k(x_t, A_t)\right).
\end{equation}
Here $\Phi_{\mathrm{local}}(\cdot)$ and $\Phi_{\mathrm{aug}}(\cdot)$ denote bounded synthesis operators that 
(i) select and compress the most relevant items, and 
(ii) format them into a single working context consumable by downstream modules, such as \textit{Inner Speech} and \textit{Query Generation}.
This makes the contribution of archival memory \emph{conditional} on the gate $s_t$, so that long-term retrieval is invoked only when needed.
We formalize archival retrieval using an embedding function $e(\cdot)$ and a similarity measure $\mathrm{sim}(\cdot,\cdot)$.
Given a query text $x_t$, the Top-$k$ retrieval operator returns
\begin{equation}
\mathrm{RETRIEVE}_k(x_t, A_t) = \operatorname{Top-}k_{a \in A_t}\; \mathrm{sim}\!\left(e(x_t), e(a)\right).
\label{eq:retrieval_topk}
\end{equation}
The retrieved set is intentionally bounded (that is, $k$ is fixed) to control context growth and to keep retrieval \textit{auditable}.
While $e(\cdot)$ and $\mathrm{sim}(\cdot,\cdot)$ are implementation-dependent, Equation~\eqref{eq:retrieval_topk} captures a key design choice, that is, archival access is mediated by semantic similarity and produces a finite set of candidate items that is subsequently filtered and synthesized into $W_t$ by $\Phi_{\mathrm{aug}}(\cdot)$.

It must emphasized that a working context must be both bounded and internally coherent to be useful for downstream control modules.
We therefore model the synthesis of a working context as producing 
(i) a bounded representation and 
(ii) a conflict-resolved set of candidate facts.
Let $\mathrm{COHERENT}(\cdot)$ denote a predicate that holds when redundancy and priority conflicts are resolved under domain-independent rules, for example, preferring the most recent explicit human-provided correction over older conflicting facts.
Then $\Phi_{\mathrm{local}}$ and $\Phi_{\mathrm{aug}}$ are required to satisfy:
\begin{equation}
\mathrm{LEN}(W_t) \le B_W
\qquad \textit{and} \qquad
\mathrm{COHERENT}(W_t)=1,
\end{equation}
for a fixed budget $B_W$, for example, \textit{token} or character budget, and a coherence measure.
Formally, $\mathrm{LEN}:\mathcal{W}\rightarrow\mathbb{N}$ maps a context $W_t \in \mathcal{W}$ to a nonnegative integer \textit{length}, and $B_W \in \mathbb{N}$ is a fixed budget expressed in the same units.
Operationally, coherence constraints are enforced before $W_t$ is returned to the controller, so that downstream modules can treat $W_t$ as a consistent evidence source.

At the end of each turn, the \textit{Memory} agent updates the persistent state through the managed operator $\mathcal{U}$ already introduced in Equation~\eqref{eq:memupdate}.
Here, we specify a concrete decomposition consistent with the hierarchical design in Figure~\ref{fig:memory_agent}.
First, new human-robot interaction traces are appended to the historic buffer, that is
\begin{equation}
H_t^{+} = \mathrm{APPEND}(H_t, \xi_t),
\end{equation}
where 
$\xi_t$ denotes the turn trace used for persistence, for example, a structured summary of $(x_t, y_t, \hat d_t, \tau_t, \theta_t,c_t, \rho_t,E_t)$.
Second, consolidation is performed only when capacity constraints are violated.
Let $H_{\max}$ be the maximum historic-buffer capacity and let $B_{\mathrm{core}}$ be the core-memory budget, that is, the character/token budget.
When $|H_t^{+}| > H_{\max}$, the oldest traces are summarized by a compression operator $\sigma(\cdot)$, and merged into core memory via a revision-aware update operator $u(\cdot)$:
\begin{equation}
|H_t^{+}| > H_{\max}\ \Rightarrow\
\begin{cases}
C_t^{+} = u\!\left(C_t,\sigma(H_t^{\mathrm{old}})\right),\\
H_{t+1} = H_t^{\mathrm{recent}}.
\end{cases}
\label{eq:hist2core}
\end{equation}
The operator $u(\cdot)$ enforces non-redundancy and resolves contradictions by applying explicit priority rules, for example, corrections override previous entries, whereas hard constraints override soft preferences.
Third, when the core-memory budget is exceeded, a split/transfer operator $\pi(\cdot)$ partitions the core content into
(i) facts retained in $C$, and 
(ii) facts transferred to the archive, such that
\begin{equation}
\mathrm{LEN}(C_t^{+}) > B_{\mathrm{core}} \ \Rightarrow\
(C_{t+1}, A_{t+1}) = \pi(C_t^{+}, A_t).
\label{eq:core2arch}
\end{equation}
The operator $\pi(\cdot)$ is constrained to preserve high-priority stable facts in $C_{t+1}$ while transferring lower-priority or rarely accessed items into $A_{t+1}$.

The output of the retrieval path is the bounded working context $W_t$, which is consumed by agents such as \textit{Inner Speech} and \textit{Query Generation}, as described in Section~\ref{sec:system_architecture}.
When external tool grounding is performed, the tool evidence $E_t$ is obtained separately.
Together, $(W_t, E_t)$ define the evidence bundle $B_t$ used by \textit{Outer Speech} and the faithfulness constraint of Equation~\eqref{eq:faithfulness}.
This separation reflects the conceptual distinction introduced in Section~\ref{sec:agent_model} between
(i) \emph{memory sufficiency}, that is, whether archival retrieval is needed to construct $W_t$, and 
(ii) \emph{tool grounding}, that is, whether external evidence $E_t$ is required to respond faithfully.

As a last remark, it should be noted that since the memory state $S_t$ is shared across domains, persistence remains available even when \textit{Scope Detection} triggers a runtime domain switch, as discussed in Section~\ref{sec:system_architecture}.
As a consequence, user identity, preferences, and long-horizon constraints can be reused when humans interacting with the robot interleave multiple topics and databases, supporting continuity across sessions and multi-domain interaction.

\section{Experimental Evaluation}
\label{sec:experimental_evaluation}

\subsection{Implementation Notes}
\label{sec:implementation_notes}

This Section summarizes concrete engineering choices adopted to realize the factored controller described in Sections~\ref{sec:agent_model}--\ref{sec:memory_model}.
The goal is to clarify both deployment and interface decisions that 
(i) enable the modular evaluation of the intermediate variables in Equations~\eqref{eq:scope}--\eqref{eq:outerspeech}, and 
(ii) provide necessary background for the experimental results reported herewith.

The implementation of \textsf{JANUS} follows a distributed design nowadays commonplace in robot cognitive and control architectures, in which all major components, that is, \textit{Perception}, the various reasoning modules, \textit{Memory}, \textit{Inner Speech}, \textit{Outer Speech}, and robot-related control modules, are deployed as ROS~2 nodes (Humble), consistently with the pipeline shown in Figure~\ref{fig:implementation}.
Communication among modules leverages publish--subscribe for streaming messages, for example, perceptual input and intermediate module outputs, and and synchronous service calls for stateful operations.
In particular, access to \textit{Memory} is exposed through ROS services to serialize reads and writes over the shared state $S_t$, which prevents \textit{race} conditions and ensuring that the working context $W_t$ used within a turn keeps consistency.

All LLM-dependent modules are orchestrated using LangChain\footnote{Link: \url{https://www.langchain.com/}} to standardize tool binding, structured outputs, and prompt management across components.
Structured representations, that is, Task Tools, \textit{Query Generation} specifications, and control outputs of \textit{Inner Speech}, are enforced through typed schemas.
This ensures that intent parameters, query specifications, and control signals are programmatically consumable by downstream modules.
A centralized configuration manifest specifies model providers and per-module parameters, such as temperature and token limits, which supports in principle heterogeneous deployments, for example, different model providers or sizes across modules.

As we discussed above, grounded execution is intentionally decoupled from the underlying knowledge base technology.
\textsf{JANUS} introduces a database-adapter layer that exposes a uniform execution interface to \textit{Query Generation}, with the aim of abstracting differences in query languages, capabilities, and result formats across heterogeneous backends.
In fact, any backend-specific logic (for example, connection handling, query execution, and result formatting) is encapsulated in adapters, and the appropriate adapter is selected at runtime based on the active domain and query type.
This design keeps \textit{Query Generation} focused on producing executable query specifications aligned with the selected task schema, while delegating execution details to adapters.
Furthermore, it enables a single domain to combine multiple backends (relational, graph, and vector stores) when needed.
In the current implementation, \textsf{JANUS} provides adapters for three database families, namely
SQLite relational databases, 
Neo4j graph databases, and 
Qdrant vector databases.
As a consequence, extending \textsf{JANUS} to new domains or data sources requires adding an adapter and registering it in the adapter registry, without modifying the core reasoning modules.

The \textit{Memory} agent is implemented as a dedicated, stateful component hosting the shared memory state $S_t = \langle H_t, C_t, A_t \rangle$, as described in Section~\ref{sec:memory_model}.
Other modules interact with it through two operations:
(i) \emph{read}, which returns the bounded working context $W_t$ for the current turn, and 
(ii) \emph{write}, which applies the managed update operator $\mathcal{U}$ at the end of the turn as provisioned by Equation~\eqref{eq:memupdate}.
The internal control structure follows the retrieval--update decomposition shown in Figure~\ref{fig:memory_agent}.
In the current implementation, these sub-steps are encoded as an explicit state machine based on LangGraph\footnote{Link: \url{https://www.langchain.com/langgraph}}, which makes the memory policy auditable and maintainable.
The archival memory $A_t$ is implemented as a vector index supporting semantic retrieval.
When the information-sufficiency gate indicates that archival retrieval is required, see  Equation~\eqref{eq:suffgate}, the agent retrieves the Top-$k$ most similar items according to an embedding function $e(\cdot)$ and a similarity measure $\mathrm{sim}(\cdot,\cdot)$, as shown in Equation~\eqref{eq:retrieval_topk}, and merges them into the synthesized working context $W_t$.
Capacity constraints are enforced through fixed budgets that directly instantiate the boundedness assumptions of Section~\ref{sec:memory_model}.
Historic memory is bounded by a maximum number of interaction traces $H_{\max}$, while core memory is bounded by a budget $B_{\mathrm{core}}$, which is measured in characters in our implementation.
As we discussed above, when $|H_t| > H_{\max}$, the oldest traces are summarized and merged into core memory, as in Equation~\eqref{eq:hist2core}.
When the core-memory budget is exceeded, the split--transfer operator $\pi(\cdot)$ partitions core content into facts to keep and facts to transfer to archival memory, according to Equation~\eqref{eq:core2arch}.
In practice, this step is applied iteratively until the budget constraint is satisfied.

The experiments reported next evaluate the \emph{intermediate variables} and interfaces introduced in the conceptual model, with a specific emphasis on domain selection $\hat d_t$, intent/schema extraction $(\tau_t, \theta_t)$, the sufficiency gate $s_t$, the control decision $c_t$, tool-grounding decisions $\rho_t$, and the \textit{groundedness} of responses under the evidence bundle $B_t = (W_t, E_t)$.
This allows us to quantify where failures arise along the factored control loop, rather than only assessing end-to-end response quality.

\begin{table}[!ht]
\centering
\scriptsize
\renewcommand{\arraystretch}{1.08}
\setlength{\tabcolsep}{4pt}
\begin{tabularx}{\linewidth}{p{0.12\linewidth} Y Y Y}

& \texttt{AddToDatabase} & \texttt{DishInfo} & \texttt{SubstituteDish} \\
\midrule
\midrule
\textbf{Description} &
Creates a new \texttt{Person} profile in the \textsc{Advisor} KG from the human message, storing nutritional targets and optional intolerance data. &
Retrieves information about a specified \texttt{Dish} and, when a user profile is provided, checks compatibility against user constraints as stored in the KG. &
Proposes an alternative \texttt{Dish} matching the user profile, and respects explicit ingredient constraints, optionally grounded in a requested day/meal context. \\

\midrule
\textbf{Required fields} &
\begin{tabular}[t]{@{}l@{}}
User identifier\\
Daily calories target\\
Daily protein target (g)\\
Daily carb target (g)\\
Daily fat target (g)
\end{tabular}
&
\begin{tabular}[t]{@{}l@{}}
Dish identifier
\end{tabular}
&
\begin{tabular}[t]{@{}l@{}}
User identifier
\end{tabular}
\\

\midrule
\textbf{Optional fields} &
\begin{tabular}[t]{@{}l@{}}
List of intolerance data
\end{tabular}
&
\begin{tabular}[t]{@{}l@{}}
User identifier\\
Ingredients to check
\end{tabular}
&
\begin{tabular}[t]{@{}l@{}}
Ingredients to exclude\\
Ingredients to prefer\\
Only-ingredients constraint\\
Day of week\\
Meal type\\
Weekly-plan flag \textit{(internal)}
\end{tabular}
\\

\midrule
\textbf{Execution notes} &
All values are extracted only from information explicitly provided by the human.
If any mandatory field is missing, \textsf{JANUS} asks for clarification rather than defaulting. &
If ingredient checks are requested, \textsf{JANUS} verifies whether the dish contains those ingredients, including via recursive ingredient composition when available in the KG. &
Ingredient constraints are interpreted as follows: 
excluded ingredients must not appear in the proposed dish; 
preferred ingredients should be included when possible; 
the only-ingredients constraint is applied \emph{only} when the assisted person explicitly states that no other ingredients are available or acceptable.
Day/meal context guides meal planning but does not override allergy or intolerance constraints. \\

\end{tabularx}
\caption{
Structured task tools used in the \textsc{Advisor} domain. 
Each column reports one task schema (Pydantic tool) that serves as the contract between \textit{Intent Recognition} and \textit{Query Generation}. 
For each tool we list: a brief description, the required parameters used by \textit{Inner Speech} to determine execution readiness, optional parameters that refine grounding and constraint checking, and a concise summary of the execution semantics that \textit{Query Generation} must implement over the dietary knowledge graph.
}
\label{tab:advisor_tools}
\end{table}

\begin{figure}[t!]
\centering
\includegraphics[width=\linewidth]{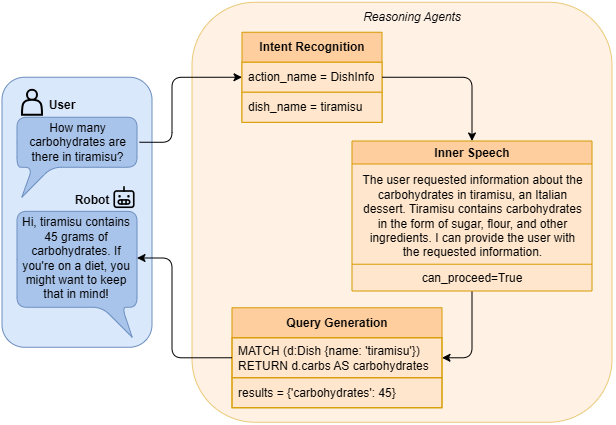}
\caption{
A complete, example interaction trace for a single turn in the \textsc{Advisor} domain.
The human asks ``How many carbohydrates are in tiramisu?''.
\textit{Intent Recognition} maps the utterance to the \texttt{DishInfo} schema $\tau_t$ and extracts a typed parameter assignment $\theta_t$ containing the dish identifier, that is, \texttt{tiramisu}.
\textit{Inner Speech} evaluates execution readiness, and issues a control decision $c_t = \textsc{Proceed}$, together with the tool-grounding decision $\rho_t = 1$.
\textit{Query Generation} produces an executable query specification $q_t$, which is executed against the domain backend to obtain evidence $E_t$ about the dish's macronutrient composition.
Then, \textit{Outer Speech} transforms the evidence bundle $B_t = (W_t, E_t)$ into an evidence-grounded natural-language reply communicating the retrieved carbohydrates value to the human.
}
\label{fig:interaction_trace}
\end{figure}

\subsection{Use Cases}
\label{sec:use_cases}

All end-to-end examples reported here are instantiated in the \textsc{Advisor} domain, a dietary-assistance scenario in which \textsf{JANUS} must interact with an assisted person over long temporal horizons to provide suggestions and insights about their diet. 
In so doing, \textsf{JANUS} maintains user-specific data, such as daily macronutrient targets and intolerance aspects, which are stored in a structured knowledge graph KG. 
The domain is \textit{intentionally} safety-relevant and interaction-heavy.
Human requests are frequently underspecified, require personalization, and must be supported by an explicit evidence $E_t$ retrieved from the KG, and incorporated in the evidence bundle $B_t = (W_t, E_t)$, as described in Section~\ref{sec:agent_model}.

The \textsc{Advisor} KG models dietary knowledge and user-specific constraints through four main entity types: 
\texttt{Person}, 
\texttt{Dish}, 
\texttt{Ingredient}, and 
\texttt{Allergen}.
A \texttt{Person} node represents a user profile and models dietary targets, for example calories and macronutrients, as well as intolerance information.
A \texttt{Dish} node is used to model meal options.
\texttt{Ingredient} nodes represent compositional elements of dishes, whereas \texttt{Allergen} nodes represent allergy categories associated with ingredients.
Edges in the KG encode both personalization and compositional structure.
A \texttt{Person} may be allergic to an \texttt{Allergen}.
A \texttt{Person} may be linked to recommended or allowed \texttt{Dish} nodes, which are used to represent diet-plan recommendations.
A \texttt{Dish} can contain an \texttt{Ingredient}, and an \texttt{Ingredient} can contain an \texttt{Allergen}.
Ingredients can recursively contain other ingredients, therefore enabling hierarchical recipes and composite foods.
This representation supports multi-hop reasoning, for example the ability to propagate allergens through ingredient composition and verify whether a dish is compatible with a specific user’s intolerance profile.

In the \textsc{Advisor} domain, user intents are mapped to three \emph{Task Tools} implemented as typed Pydantic schemas.
These task schemas instantiate the abstract library $\mathcal{T}(d)$ introduced in Section~\ref{sec:agent_model}.
They define both the task semantics and the typed parameter space used by downstream modules.
In particular, they determine which fields \textit{Inner Speech} treats as mandatory for execution readiness, as foreseen in Equation~\eqref{eq:comp}, and which optional fields refine constraint checking and retrieval.
Table~\ref{tab:advisor_tools} summarizes the three tools, their required and optional fields, as well as their execution semantics over the \textsc{Advisor} KG.
In particular:
\begin{itemize}
\item 
\texttt{AddToDatabase} creates a new \texttt{Person} entry in the KG starting from the human input, which stores their identifier and daily nutritional targets.
Intolerance data are stored only when explicitly provided.

\item
\texttt{DishInfo} retrieves information about a specified \texttt{Dish} and, when a user profile is available, evaluates compatibility against their constraints stored in the KG. 
The request may also include ingredient-level checks.

\item
\texttt{SubstituteDish} proposes an alternative \texttt{Dish} that is compatible with both user profile and explicit ingredient constraints. 
The request may specify ingredients to exclude, ingredients to prefer, or an \emph{only-ingredients} constraint that applies when the human explicitly restricts the available ingredients.
Optional day/meal context supports meal planning but does not override allergy or intolerance constraints.
\end{itemize}

The structured task definitions serve as the \textit{contract} between \textit{Intent Recognition} and \textit{Query Generation}.
They expose the intermediate variables evaluated in Section~\ref{sec:experimental_evaluation}, that is, schema selection $\tau_t$, parameter assignment $\theta_t$, readiness and clarification decisions via $c_t$, and grounded evidence retrieval via $(q_t,E_t)$.
The examples reported next illustrate representative requests for each task type, including cases in which \textit{Inner Speech} triggers clarification when mandatory information is missing, and cases whereby query results are surfaced through \textit{Outer Speech} as evidence-backed explanations.

\subsection{Validation Metrics}
\label{sec:validation}

The results reported in this Section are obtained from a suite of module-level \emph{unit tests} in which each agent is evaluated \emph{in isolation} from the rest of the pipeline.
This protocol is aligned with the factored-control view introduced in Section \ref{sec:agent_model}.
Each unit test targets one intermediate variable, for example, $\hat d_t$, $(\tau_t, \theta_t)$, $c_t$, $\rho_t$, $q_t$, and $E_t$, and verifies whether the corresponding module behaves according to the typed contract and the intended semantics.
For each agent, we built a curated test set consisting of structured inputs and a reference output, and we traced executions using LangSmith\footnote{Link: \url{https://smith.langchain.com/}}.
Each test item therefore specifies
(i) what the agent receives at runtime, including the relevant context bundle, and 
(ii) the expected behavior of that agent under the same conditions.

\paragraph{Configuration}
All unit tests were run via the Groq API using {\small\texttt{meta\allowbreak-llama\allowbreak/llama-4\allowbreak-scout-17b\allowbreak-16e\allowbreak-instruct}}.
The sampling temperature $T$ was set per module according to its role in the pipeline.
For \textit{Scope Detection}, \textit{Intent Recognition}, and \textit{Query Generation} we use $T = 0$ to favor deterministic, reproducible structured outputs, since these modules synthesize control-relevant decisions, and variability can directly propagate into downstream errors.
\textit{Inner Speech} is run at $T = 0.2$ to allow for limited flexibility in self-monitoring, and in identifying missing or inconsistent parameters while preserving stability of execution-readiness decisions.
Instead, \textit{Outer Speech} is run at $T = 0.4$ to improve the naturalness and readability of explanations for humans.
This module does not affect grounding execution, and moderate linguistic diversity is acceptable as long as it remains faithful to the provided evidence bundle.
Additionally, all tests were carried out on a laptop equipped with an Intel Core i7 7700HQ, an Nvidia GTX 1050Ti with 4GB of VRAM and 32GB of RAM.

\begin{table}[t!]
\centering
\scriptsize
\begin{tabularx}{\textwidth}{c l Y Y}
\textbf{Id} & \textbf{Domain} & \textbf{Description} & \textbf{Supported tasks} \\
\midrule
\midrule
1 & 
\textsc{Movies} &
Explore a movie and cinema schedule database. &
\texttt{MovieInfo}, \texttt{TimetableInfo} \\
\midrule
2 & 
\textsc{Advisor} &
Support the assisted person in following a proper dietary regimen. &
\texttt{AddToDatabase}, \texttt{DishInfo}, \texttt{SubstituteDish} \\
\midrule
3 & 
\textsc{Travel} &
Plan trips and assist with travel reservations and advice. &
\texttt{FlightSearch}, \texttt{HotelBooking}, \texttt{TravelInfo} \\
\midrule
4 & 
\textsc{JobHunt} &
Find jobs and manage applications. &
\texttt{JobSearch}, \texttt{ApplicationStatus}, \texttt{ResumeReview} \\
\midrule
5 &
\textsc{TechSupport} &
Provide technical support for devices and software. &
\texttt{TroubleshootDevice}, \texttt{SoftwareHelp}, \texttt{WarrantyInfo} \\
\midrule
6 & 
\textsc{Education} &
Help students organize and understand their academic journey. &
\texttt{CourseInfo}, \texttt{ExamSchedule}, \texttt{AssignmentHelp} \\
\midrule
7 & 
\textsc{HealthSymptoms} &
Support the assisted person in interpreting symptoms and medication questions. &
\texttt{SymptomCheck}, \texttt{MedicationInfo}, \texttt{DoctorReferral} \\
\midrule
8 & 
\textsc{Shopping} &
Find and compare products and track orders. &
\texttt{ProductSearch}, \texttt{CompareProducts}, \texttt{TrackOrder} \\
\midrule
9 & 
\textsc{Events} &
Discover local events and support ticketing and reminders. &
\texttt{EventSearch}, \texttt{TicketBooking}, \texttt{EventReminder} \\
\midrule
10 & 
\textsc{Restaurants} &
Discover restaurants and support menu questions and bookings. &
\texttt{RestaurantSearch}, \texttt{MenuInfo}, \texttt{TableBooking} \\
\midrule
11 & 
\textsc{Fitness} &
Support the assisted humans with workout planning, exercise info, and progress logging. &
\texttt{WorkoutPlan}, \texttt{ExerciseInfo}, \texttt{ProgressTracking} \\
\midrule
12 & 
\texttt{OutOfScope} &
Fallback domain selected when no other domain matches the request. &
-- \\
\end{tabularx}
\caption{
The table shows the domains used to test the \textit{Scope Detection} agent. 
For each domain, we present the textual description and the list of supported tasks that were passed as context to agents to ground their decision.
}
\label{tab:scope_domains}
\end{table}

\paragraph{Scope Detection}
Each \textit{Scope Detection} test item contains a request from the assisted person, and the list of candidate domains provided by the \textit{Customization Layer}.
The reference output specifies the correct domain to activate (or \texttt{OutOfScope} when none applies), together with a short justification.
Although the end-to-end examples in this paper focus on the \textsc{Advisor} domain, scope detection must operate under multi-domain ambiguity.
Therefore, we defined the 11 domain profiles shown in Table~\ref{tab:scope_domains}.

\begin{figure}[t!]
\centering
\includegraphics[width=\linewidth]{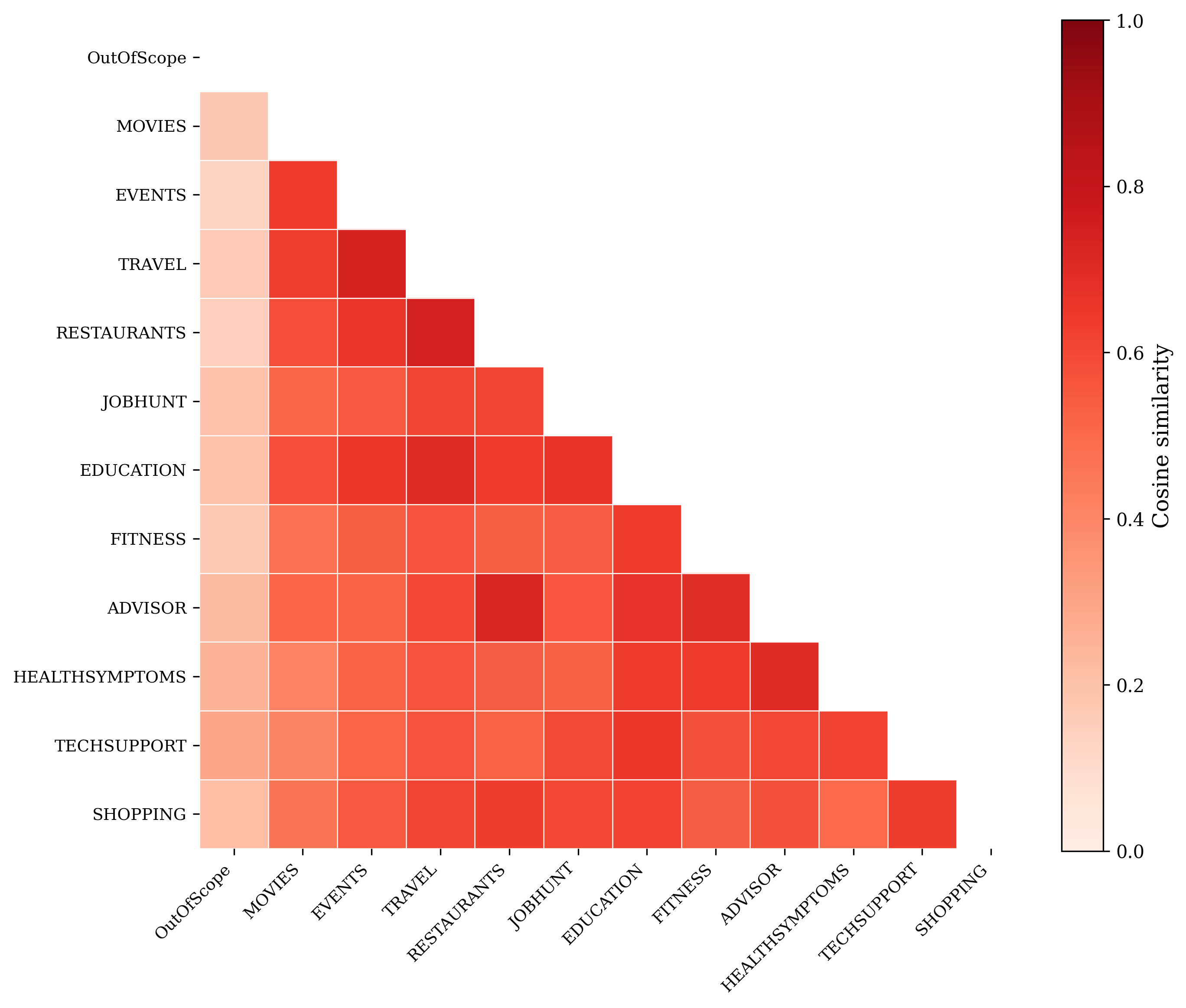}
\caption{
Inter-domain semantic similarity used in the \textit{Scope Detection} study. 
The heatmap reports the pairwise cosine similarity between the textual domain descriptions in Table~\ref{tab:scope_domains}, including the \texttt{OutOfScope} fallback. 
Darker cells indicate higher similarity, that is, greater semantic overlap, whereas lighter cells indicate lower similarity. 
The matrix is shown in triangular form for the sake of readability.}
\label{fig:domain_similarity}
\end{figure}

\begin{table}[t!]
\centering
\scriptsize
\begin{tabularx}{\linewidth}{l Y r}
\textbf{Code} & \textbf{Included Domains} & \textbf{Avg.\ cosine sim.} \\
\midrule
\midrule
\textsc{MHS} & \textsc{Movies}, \textsc{HealthSymptoms}, \textsc{Shopping}, \texttt{OutOfScope} & 0.461 \\
\midrule
\textsc{HSR} & \textsc{HealthSymptoms}, \textsc{Shopping}, \textsc{Restaurants}, \texttt{OutOfScope} & 0.557 \\
\midrule
\textsc{ATT} & \textsc{Advisor}, \textsc{Travel}, \textsc{TechSupport}, \texttt{OutOfScope} & 0.587 \\
\midrule
\textsc{TEF} & \textsc{TechSupport}, \textsc{Education}, \textsc{Fitness}, \texttt{OutOfScope} & 0.623 \\
\midrule
\textsc{TER} & \textsc{Travel}, \textsc{Events}, \textsc{Restaurants}, \texttt{OutOfScope} & 0.715 \\
\midrule
\texttt{ALL} & All domains in Table~\ref{tab:scope_domains} & -- \\
\end{tabularx}
\caption{
Domain configurations for \textit{Scope Detection} used to study the effect of inter-domain semantic similarity.
The average cosine similarity is computed over the three pairwise similarities within each $3$-domain set.
The full-set configuration \texttt{ALL} is evaluated on a separate curated set of 66 items.
}
\label{tab:scope_combinations}
\end{table}

\textit{Scope detection} is evaluated in two stages. 
First, we run a full-set test using the configuration \texttt{ALL}, which includes all $11$ domains of Table~\ref{tab:scope_combinations} and \texttt{OutOfScope}, on a curated set of $N=66$ labeled requests.
Second, we investigate the effect of inter-domain semantic similarity on classification difficulty. 
We embed each textual domain description using a sentence-transformer encoder, and compute a cosine similarity matrix between all domain pairs, which is reported in Figure~\ref{fig:domain_similarity}. 
For any $3$-domain candidate set \(\{a, b, c\}\), we define its average similarity as
\[
\mathrm{sim}_{\mathrm{avg}}(a,b,c) \;=\; \frac{s(a,b)+s(a,c)+s(b,c)}{3},
\]
where \(s(\cdot,\cdot)\) denotes cosine similarity between the corresponding domain description embeddings. 
We then select five $3$-domain configurations that span increasing values of \(\mathrm{sim}_{\mathrm{avg}}\), and re-run \textit{Scope Detection} unit tests using each configuration as the candidate set, always including the same \texttt{OutOfScope} fallback. 
For compact reference, we denote these configurations by short codes formed by concatenating domain initials as shown in Table~\ref{tab:scope_combinations}.
The selected $3$-domain configurations cover distinct overlap regimes in Figure~\ref{fig:domain_similarity}. 
\textsc{MHS} targets a low-overlap setting, in which the three domains are comparatively distinct in the similarity matrix. 
\textsc{TER} is related to a high-overlap setting, where \textsc{Travel}, \textsc{Events}, and \textsc{Restaurants} form a tight cluster with strong mutual connections. 
The remaining configurations, namely \textsc{HSR}, \textsc{ATT}, and \textsc{TEF} refer to intermediate overlap levels. 
Each configuration is evaluated on $24$ test items, which enables a comparison of scope-detection behavior as the candidate-domain set shifts from semantically distinct to strongly overlapping.

\paragraph{Intent Recognition}
Each \textit{Intent Recognition} test item contains a request from the assisted person, and the set of supported tasks for the active domain, including their typed schemas.
The reference output specifies 
(i) the correct task/tool to activate, and 
(ii) the typed parameter values to extract from the request.
This test isolates the semantic parsing step that converts free-form language into the structured task representation $(\tau_t, \theta_t)$.
\textit{Intent Recognition} is tested on $N=51$ test items.

Parameter extraction is evaluated by comparing the predicted argument dictionary $A$ against the reference dictionary $E$.
Let the true positives be
\[
TP \;=\; \sum_{k \in \mathrm{keys}(E)} \mathbb{I}\!\left[A(k)=E(k)\right],
\]
where $\mathbb{I}[\cdot]$ is the indicator function and $A(k)$ is undefined if $k \notin \mathrm{keys}(A)$.
Precision and recall are computed as
\[
P \;=\; 
\begin{cases}
\frac{TP}{|\mathrm{keys}(A)|} & \text{if } |\mathrm{keys}(A)|>0,\\
0 & \text{otherwise,}
\end{cases}
\qquad
R \;=\;
\begin{cases}
\frac{TP}{|\mathrm{keys}(E)|} & \text{if } |\mathrm{keys}(E)|>0,\\
0 & \text{otherwise.}
\end{cases}
\]
Finally, the parameter-extraction F1 score is
\[
F1 \;=\;
\begin{cases}
\frac{2PR}{P+R} & \text{if } P+R>0,\\
0 & \text{otherwise.}
\end{cases}
\]
In this paper, we report $TP$, precision, recall, and the F1 score because parameter extraction is naturally a \emph{set-valued} prediction problem over typed slots, that is, the model must both 
(i) identify \emph{which} fields are present, and 
(ii) assign \emph{correct values}.
In this setting, precision penalizes spurious or hallucinated fields in $A$ (over-extraction), recall penalizes missing required information (under-extraction), and the F1 score summarizes the trade-off into a single scalar suitable for comparing configurations.
This is particularly relevant for \textsf{JANUS}, because errors in either direction propagate to downstream control, for example, \textit{Inner Speech} readiness and clarification decisions, and to grounded execution, that is, \textit{Query Generation}.
Therefore, we require a metric that separately exposes omission \textit{versus} commission errors rather than reporting only task-selection accuracy.

\paragraph{Inner Speech}
Each \textit{Inner Speech} test item contains the request by the assisted person, and the structured output produced by \textit{Intent Recognition}.
The reference output specifies an \emph{execution-readiness} label, that is, whether the request can be fulfilled given the extracted parameters, and a reference \textit{Inner Speech} rationale explaining \textit{why} the current information is sufficient to proceed or, conversely, which missing or inconsistent elements require clarification.
This unit test isolates the control role of \textit{Inner Speech} as a \emph{parameter-completeness and consistency gate}, that is, preventing under-specified grounding and triggering an information-seeking turn when required.
\textit{Inner Speech} is tested on $N = 40$ items.

\paragraph{Query Generation}
Each \textit{Query Generation} test item contains the request from the human, the structured intent output, that is, the assigned task and extracted parameters, and the assumed domain data source.
In the experiments we report here, grounding is performed against a knowledge graph.
Therefore the reference outputs include a target \emph{Cypher} query (or queries) and the corresponding expected results.
This test isolates the ability of \textit{Query Generation} to produce executable, domain-grounded retrieval steps $q_t$ aligned with the intended information need, and the typed schema constraints.
\textit{Query Generation} is tested on $N = 30$ items.

We evaluate query grounding using two complementary measures.
The \emph{query validity rate} is the fraction of generated Cypher queries that execute successfully against the Neo4j backend.
The \emph{result overlap} score compares the results returned by the generated query against the curated reference results.
Since Neo4j outputs may contain nested lists and structured records, we apply a deterministic normalization step that flattens each result into a collection of comparable items.
Structured records are kept as dictionaries, while all other elements are treated as atomic values.
Ordering differences are ignored.
If the generated execution contains an explicit query failure marker, the overlap score is defined as $0$.

Let $R$ be the reference result and $G$ the generated result for a test item.
After normalization, each result is represented as a pair of collections, namely dictionary items and atomic items, denoted $(D_R, A_R)$ and $(D_G, A_G)$, respectively.
For a dictionary $d$, let $\mathrm{Keys}(d)$ be its set of keys, and let
\[
\mathrm{Items}(d) \;=\; \{(k, d[k]) \;:\; k \in \mathrm{Keys}(d)\}
\]
be the set of its key--value pairs.
We say that $d_2$ \emph{covers} $d_1$ if all key--value pairs of $d_1$ are contained in $d_2$, that is,
\[
\mathrm{Items}(d_1)\subseteq \mathrm{Items}(d_2).
\]
A generated dictionary $g\in D_G$ matches a reference dictionary $r\in D_R$ if either $\mathrm{Items}(g) \subseteq \mathrm{Items}(r)$ or $\mathrm{Items}(r) \subseteq \mathrm{Items}(g)$.
Let $M$ be the number of one-to-one dictionary matches obtained by greedily matching each $g\in D_G$ to at most one unused $r \in D_R$ under this criterion.
Atomic items are compared by set overlap:
\[
I_A = |A_R \cap A_G|,\qquad U_A = |A_R \cup A_G|.
\]
The dictionary-union size is computed as
\[
U_D = |D_R| + |D_G| - M.
\]
If $U_D + U_A = 0$, overlap is defined as $1$.
Otherwise, the final result-overlap score is
\[
\mathrm{overlap}(R,G) \;=\; \frac{M + I_A}{U_D + U_A}.
\label{eq:overlap}
\]

\paragraph{Outer Speech}
Each \textit{Outer Speech} test item contains: user request, the generated query/queries, the retrieved results.
The reference output is a target explanation that provides an answer to the assisted person, and one that justifies the answer using the provided evidence, making clear what was retrieved and how it supports the response.
This unit test isolates the role of \textit{Outer Speech} in transforming the evidence bundle $B_t = (W_t, E_t)$ into a faithful explanation for the human satisfying the faithfulness constraint of Eq.~\ref{eq:faithfulness}.
Outer Speech is tested on $N = 31$ items.

\begin{table}[t!]
\centering
\scriptsize
\begin{tabular}{lll}
\textbf{Module} & \textbf{Metric} & \textbf{Value} \\
\midrule
\midrule
Scope detection & Domain selection accuracy (\textsc{MHS}) & 100.00\% \\
 & Domain selection accuracy (\textsc{HSR}) & 95.80\% \\
 & Domain selection accuracy (\textsc{ATT}) & 100.00\% \\
 & Domain selection accuracy (\textsc{TEF}) & 100.00\% \\
 & Domain selection accuracy (\textsc{TER}) & 100.00\% \\
 & Domain selection accuracy (\texttt{ALL}) & 98.50\% \\
\midrule
Intent recognition & Task selection accuracy & 100.00\% \\
 & Parameter extraction F1 & 1.00 \\
\midrule
Inner speech & Execution-readiness accuracy & 100.00\% \\
\midrule
Query generation & Query validity rate & 100.00\% \\
 & Result overlap (\textit{versus} reference) & 96.70\% \\
\end{tabular}
\caption{
Agreement metrics from isolated unit tests of the main agents. 
\textit{Scope Detection} accuracy is reported for the full candidate set (\texttt{ALL} 11 domains \textit{plus} \texttt{OutOfScope}, that is, $66$ items), and for the five $3$-domain configurations (\textsc{MHS}, \textsc{HSR}, \textsc{ATT}, \textsc{TEF}, \textsc{TER}, $24$ items each). 
\textit{Intent Recognition} evaluates task selection and typed parameter extraction against curated references, 
\textit{Inner Speech} evaluates the execution-readiness decision that gates whether grounding can proceed without clarification, whereas
\textit{Query Generation} evaluates query executability, and the overlap between retrieved results and reference outcomes.
}
\label{tab:module-metrics}
\end{table}

\begin{figure}[!t]
\centering
\includegraphics[width=\linewidth]{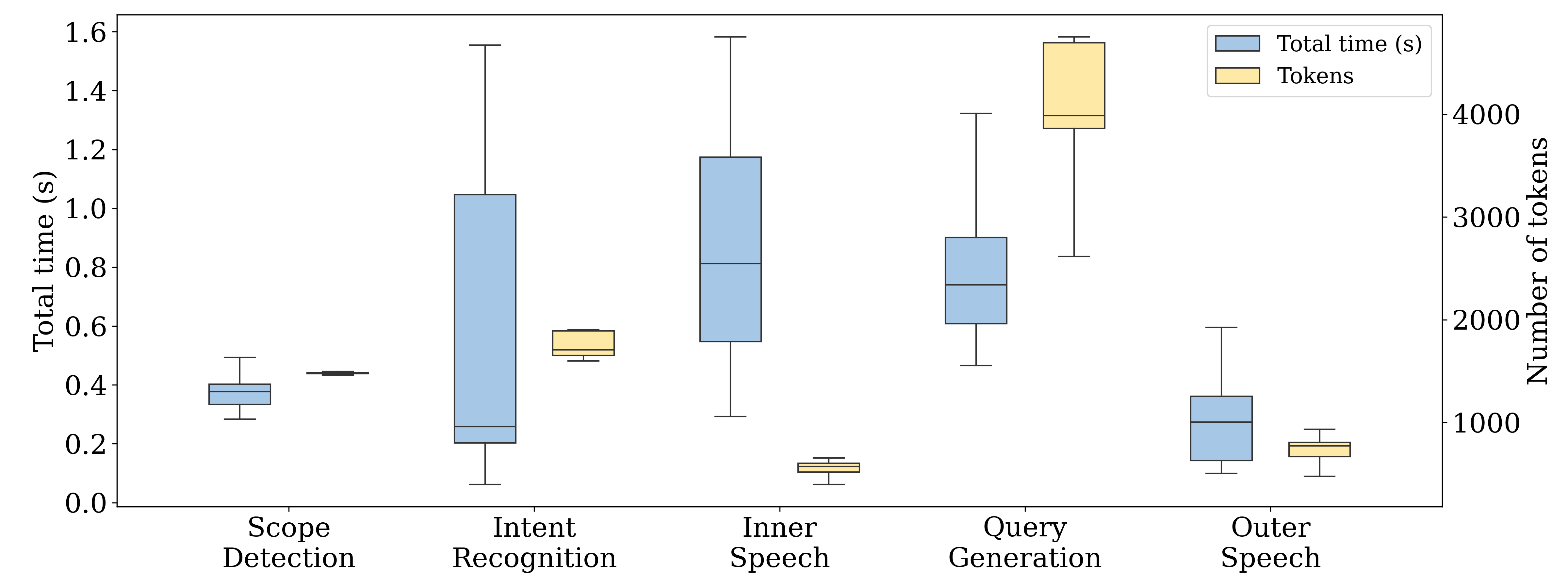}
\caption{
Distribution of \textit{total processing time}, shown on the left y-axis, and \textit{number of generated tokens}, shown on the right y-axis, for the main modules of \textsf{JANUS}.}
\label{fig:boxplot}
\end{figure}

\subsection{Validation Results}
\label{sec:validation}

We report here three complementary views of the overall behavior of \textsf{JANUS}.
First, we quantify agent-level correctness by agreement with curated references (Table~\ref{tab:module-metrics}), which corresponds to the correctness of the intermediate variables in the factored controller.
Second, we report efficiency statistics in terms of latency and token usage (Figure~\ref{fig:boxplot}) to characterize responsiveness at run-time.
Third, we evaluate the perceived quality of the textual fragments produced by the text-emitting agents, that is, \textit{Scope Detection}, \textit{Inner Speech}, and \textit{Outer Speech} via a mixed protocol combining human ratings and automatic judging based on G-Eval~\cite{liu2023g}.

\paragraph{Quantitative agent-level correctness}
Table~\ref{tab:module-metrics} reports agreement metrics obtained from the isolated unit tests.
It is possible to observe that 
\textit{Scope Detection} achieves 98.50\% accuracy in the full candidate setting (\texttt{ALL}), indicating that domain routing remains reliable even when the candidate set includes $11$ domains \textit{plus} \texttt{OutOfScope}.
In the $3$-domain study, accuracy is \textit{perfect} in four configurations (\textsc{MHS}, \textsc{ATT}, \textsc{TEF} and \textsc{TER}) and slightly lower in \textsc{HSR} (95.80\%).
It is noteworthy that the only observed drop does not occur in the highest-overlap triad (that is, \textsc{TER}), suggesting that at the present scale the model is robust to moderate semantic entanglement in domain descriptions, although stronger effects may emerge with larger candidate sets or finer-grained domain definitions.
\textit{Intent Recognition} reaches 100.00\% task-selection accuracy and a parameter-extraction F1 of 1.00, indicating stable mapping from free-form language to typed task schemas.
\textit{Inner Speech} attains 100.00\% execution-readiness accuracy, correctly gating whether \textsf{JANUS} should proceed with grounding or ask for clarification when mandatory slots are missing.
\textit{Query Generation} achieves 100.00\% query validity, meaning that all generated Cypher queries execute successfully against Neo4j, and a result-overlap score of 96.70\%, indicating that most generated queries retrieve evidence consistent with the curated reference results.
The residual gap in the overlap metric of Equation~\ref{eq:overlap} is consistent with near-miss cases such as alternative but semantically equivalent query patterns, partial record coverage, or differences in aggregation/granularity that still yield executable queries.

\paragraph{Efficiency and computational profiles}
Figure~\ref{fig:boxplot} summarizes the distribution of processing time and generated tokens per module.
Median latencies are 0.37\,s for \textit{Scope Detection}, 0.26\,s for \textit{Intent Recognition}, 0.81\,s for \textit{Inner Speech}, 0.74\,s for \textit{Query Generation}, and 0.27\,s for \textit{Outer Speech}.
It is possible to observe that two trends stand out.
First, \textit{Query Generation} exhibits the highest token usage \textit{and} the largest latency dispersion, which is expected given that it must satisfy strict formatting constraints while producing executable code-like artifacts (and, in complex cases, multi-step queries).
Second, \textit{Inner Speech} has relatively higher median latency despite lower token usage than \textit{Query Generation}.
This reflects its deliberative role (in terms of slot validation, consistency checks, and control decisions) rather than long-form generation.
If we use per-module median latencies as a coarse estimate, turn-level response times vary with the interaction outcome.
A pure domain-routing response (that is, \textit{Scope Detection} alone) yields $\approx$0.37\,s delay.
A clarification turn (that is, \textit{Scope Detection} and \textit{Intent Recognition} and \textit{Inner Speech} and \textit{Outer Speech}) yields $\approx$1.72\,s.
A single evidence-backed answer (including \textit{Query Generation} and \textit{Outer Speech}) yields $\approx$2.46\,s, which excludes additional deployment overheads such as speech input/output, network transport, and robot actuation.
These estimates highlight that the dominant contributors to responsiveness are LLM-based inference and prompt length, rather than any single non-LLM architectural step.
Token usage mirrors this profile, with \textit{Query Generation} being the most token-consuming agent with a median of 3986 tokens, followed by \textit{Intent Recognition} with 1705 tokens, \textit{Scope Detection} with 1477 tokens, \textit{Outer Speech} with 772 tokens, and \textit{Inner Speech} with 572 tokens.

\paragraph{Qualitative evaluation of textual outputs}
Beyond module-level correctness, we evaluate the \emph{perceived} quality of natural-language fragments produced by the three text-emitting agents in \textsf{JANUS}, namely \textit{Scope Detection}, \textit{Inner Speech}, and \textit{Outer Speech}.
This analysis complements the unit-test metrics by probing the architectural principle introduced in Section~\ref{sec:agent_model}, that is, \textsf{JANUS} separates \emph{control-oriented} text used to govern the interaction flow from \textit{robot-to-human} communication, which is optimized for clarity and trust calibration.

We adopt a mixed protocol that combines human ratings and automatic judging, based on G-Eval.
We construct a balanced qualitative set by sampling $30$ items from the validation set of each agent, for a total of $90$ items.
To mitigate evaluator fatigue and reduce ordering effects, we created three evaluation forms, each containing $30$ items uniformly sampled across agents, that is, $10$ from \textit{Scope Detection}, $10$ from \textit{Inner Speech}, and $10$ from \textit{Outer Speech}), therefore covering the full $90$-item pool.
Each form is completed by $6$ participants with heterogeneous backgrounds, which is expected to reduce bias toward a single expertise profile.

For each item, evaluators are shown 
(i) the original human request, 
(ii) the agent-specific context bundle provided to \textsf{JANUS} at that turn, and 
(iii) the textual fragment produced by the evaluated agent.
Evaluators are asked to assess whether the fragment is easy to understand, remains anchored to the provided context (rather than introducing unsupported content), and helps interpret why \textsf{JANUS} behaved as it did.
Ratings are collected on a 1--5 Likert scale along three criteria:

\begin{enumerate}
\item[C1)]
\textit{Clarity}:
degree to which the fragment is easy to read and unambiguous.
\item[C2)]
\textit{Contextual relevance}:
degree to which the fragment is consistent with the provided context, and does not introduce unrelated or unsupported content.
\item[C3)]
\textit{Behavioral interpretability}:
degree to which the fragment helps a reader understand the behavior of \textsf{JANUS}, specifying what decision or action was taken and why, given the available information.
\end{enumerate}

The agent-specific context bundle matches the contracts defined in Section~\ref{sec:agent_model} and in Equations~\eqref{eq:scope}--\eqref{eq:outerspeech}.
For \textit{Scope Detection}, the context bundle consists of the candidate domains exposed by the Customization Layer.
In the case of \textit{Inner Speech}, it consists of the selected task schema and the extracted typed parameters.
For \textit{Outer Speech}, it is related to the generated query specification(s) and, when applicable, the retrieved tool evidence. 
The same rubric was applied to both human participants and automatic judging using a 5-point Likert scale, and is reported below. 
The provided text is as follows.

\begin{quote}
\small{
\textit{
``The fragment shown below is a short text produced by the system during an interaction step, generated by one of three agents: Scope Detection, Inner Speech, or Outer Speech.}

\textit{Evaluate whether it is understandable and whether it faithfully reflects the provided context, without relying on unsupported assumptions, and whether it helps interpret the system’s behavior.}

\textit{Judge each criterion using a 5-point Likert scale}

\textit{(1 = very poor, 5 = excellent).}

\textit{Criteria:}

\textit{(1) Clarity: How clear is this explanation?}

\textit{(2) Contextual relevance: How relevant is this explanation to the given context (user request, query, and results)?}

\textit{(3) Behavioral interpretability: How useful is this explanation for understanding the system's behavior?''}}
\end{quote}

Automatic judging is performed by running the same rubric through G-Eval using a set of contemporary judge models, namely Kimi K2 (Groq), Llama Versatile (Groq), Llama Scout (Groq), DeepSeek Chat, DeepSeek Reasoner, GPT-5 mini, and GPT-5.1 mini-codex.
These models are selected to span multiple providers and model families, including instruction-tuned and reasoning-oriented variants, in order to probe the robustness of automatic scoring across heterogeneous evaluators.

\begin{figure}[!t]
\centering
\includegraphics[width=\linewidth]{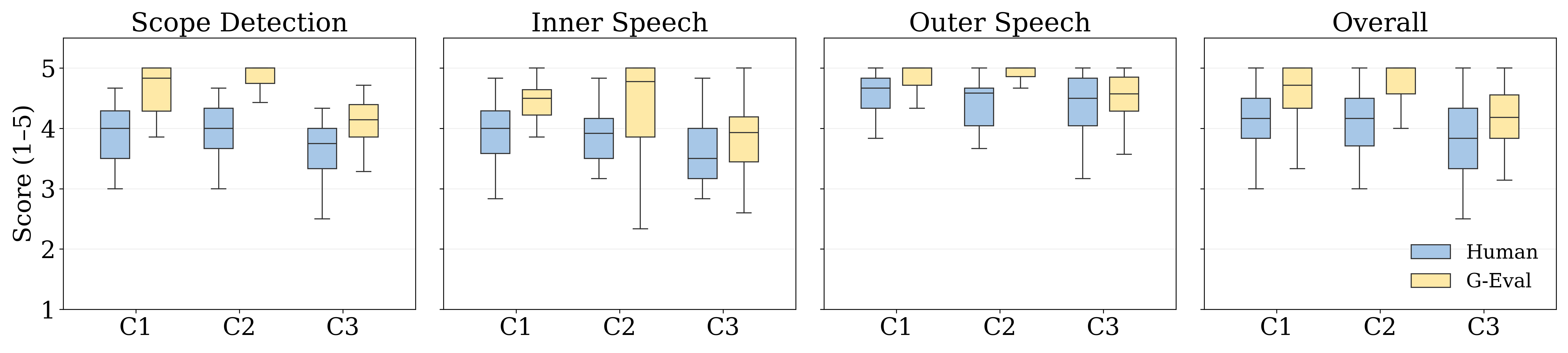}
\caption{
Distribution of qualitative ratings collected from human participants and automatic judging with G-Eval on a 5-point Likert scale.
Boxplots are shown for each agent and for the aggregation \textit{across} agents, separately for the three criteria clarity (C1), contextual relevance (C2), and behavioral interpretability (C3).
For each criterion, blue and yellow boxplots report human and G-Eval scores, respectively.}
\label{fig:form-boxplots}
\end{figure}

\begin{figure}[!t]
\centering
\includegraphics[width=\linewidth]{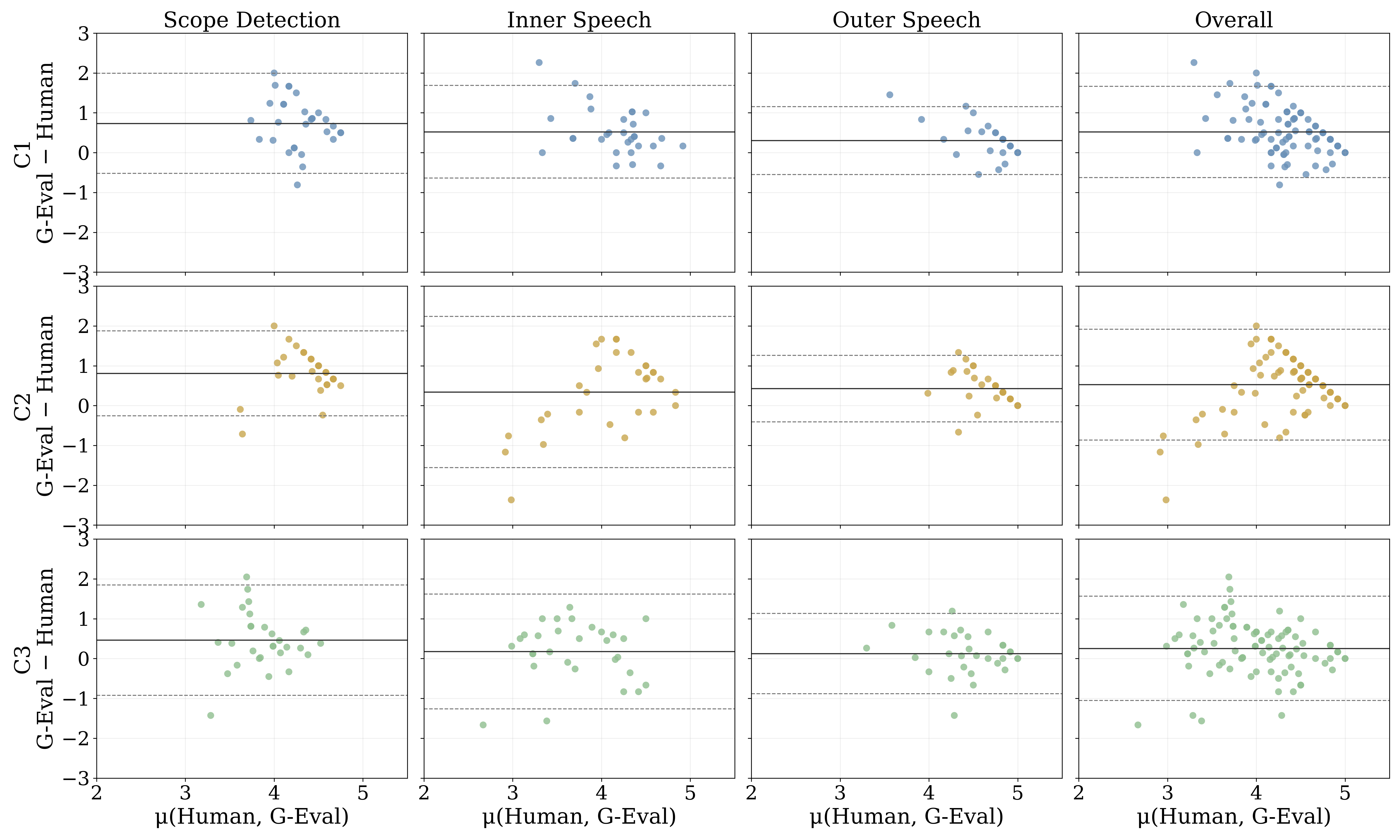}
\caption{
Bland--Altman plots comparing human rating against automatic ratings from G-Eval for the three qualitative criteria clarity (C1), contextual relevance (C2), and behavioral interpretability (C3).
Columns report results for \textit{Scope Detection}, \textit{Inner Speech}, \textit{Outer Speech}, and the \textit{aggregation} across all agents.
Each point corresponds to one evaluated item.
The x-axis reports the mean of the human and G-Eval scores, whereas the y-axis reports their difference (G-Eval \textit{minus} Human).
The solid black horizontal line denotes the mean difference, whereas black dashed lines denote the 95\% limits of agreement.
}
\label{fig:bland-altman}
\end{figure}

Figure~\ref{fig:form-boxplots} reports the distribution of qualitative ratings (both per agent and aggregated), whereas Table~\ref{tab:qualitative_results_vertical} reports the corresponding mean scores and mean absolute errors (MAE) between human and automatic ratings.
Across agents, \textit{Outer Speech} receives the highest human ratings, which is consistent with its role as the dedicated communication policy conditioned on the evidence bundle $B_t$ as of Equation~\eqref{eq:outerspeech}.
In contrast, \textit{Scope Detection} and \textit{Inner Speech} represent control-oriented traces, which are optimized for routing and execution-readiness decisions, and are therefore expected to be less ``explanatory'' when evaluated as standalone text fragments.
Pooling ratings across agents, rank correlation between human and G-Eval scores is the strongest for C3 ($\rho = 0.46$, $p = 4.55 \times 10^{-6}$), followed by C1 ($\rho = 0.36$, $p = 4.55 \times 10^{-4}$), and finally C2 ($\rho = 0.27$, $p = 1.04 \times 10^{-2}$).
This suggests that automatic judging tracks human ordering best when assessing behavioral interpretability, whereas agreement is weaker for contextual relevance.

It is possible to observe that human ratings follow the intended functional separation of \textsf{JANUS}.
\textit{Outer Speech} is consistently \textit{perceived} as the clearest and most useful textual channel, while \textit{Inner Speech} and \textit{Scope Detection} are judged less readable and less helpful when assessed as standalone textual fragments.
Across all agents, G-Eval systematically \emph{overestimates} perceived quality with respect to human perception, and the gap increases as outputs become more schematic and decision-oriented.
This suggests that LLM-based judging is comparatively reliable on robot-to-human explanations, but less trustworthy for evaluating internal control textual flow, where human usefulness depends on whether the fragment truly supports interpretation under the given context.

\begin{table}[t]
\centering
\scriptsize
\renewcommand{\arraystretch}{1.12}
\setlength{\tabcolsep}{4pt}
\begin{tabular}{lllc c}
\textbf{Agent} & \textbf{Criterion} & \textbf{Rater} & \textbf{Score} - Mean (SD) & \textbf{MAE} \\
\midrule
\midrule
\multirow{6}{*}{\makecell{Outer\\Speech}}
& \multirow{2}{*}{C1 - Clarity}
& Human  & 4.53 (SD=0.50) & \multirow{2}{*}{0.39} \\
& & G-Eval & 4.83 (SD=0.27) & \\
\cmidrule(lr){2-5}
& \multirow{2}{*}{C2 - Contextual relevance}
& Human  & 4.44 (SD=0.40) & \multirow{2}{*}{0.49} \\
& & G-Eval & 4.87 (SD=0.26) & \\
\cmidrule(lr){2-5}
& \multirow{2}{*}{C3 - Behavioral interpretability}
& Human  & 4.37 (SD=0.52) & \multirow{2}{*}{0.39} \\
& & G-Eval & 4.50 (SD=0.45) & \\
\midrule
\multirow{6}{*}{\makecell{Inner\\Speech}}
& \multirow{2}{*}{C1 - Clarity}
& Human  & 3.89 (SD=0.59) & \multirow{2}{*}{0.59} \\
& & G-Eval & 4.42 (SD=0.38) & \\
\cmidrule(lr){2-5}
& \multirow{2}{*}{C2 - Contextual relevance}
& Human  & 3.91 (SD=0.49) & \multirow{2}{*}{0.85} \\
& & G-Eval & 4.25 (SD=0.92) & \\
\cmidrule(lr){2-5}
& \multirow{2}{*}{C3 - Behavioral interpretability}
& Human  & 3.61 (SD=0.61) & \multirow{2}{*}{0.61} \\
& & G-Eval & 3.79 (SD=0.63) & \\
\midrule
\multirow{6}{*}{\makecell{Scope\\Detection}}
& \multirow{2}{*}{C1 - Clarity}
& Human  & 3.91 (SD=0.46) & \multirow{2}{*}{0.82} \\
& & G-Eval & 4.65 (SD=0.37) & \\
\cmidrule(lr){2-5}
& \multirow{2}{*}{C2 - Contextual relevance}
& Human  & 3.98 (SD=0.39) & \multirow{2}{*}{0.88} \\
& & G-Eval & 4.79 (SD=0.41) & \\
\cmidrule(lr){2-5}
& \multirow{2}{*}{C3 - Behavioral interpretability}
& Human  & 3.63 (SD=0.50) & \multirow{2}{*}{0.65} \\
& & G-Eval & 4.10 (SD=0.47) & \\
\end{tabular}
\caption{
Human and G-Eval ratings of the textual fragments produced by \textit{Scope Detection}, \textit{Inner Speech}, and \textit{Outer Speech}.
For each agent, $30$ items were evaluated with a 1--5 Likert rubric along the three criteria C1, C2, and C3.
Scores are reported as mean and standard deviation.
MAE is computed between human and G-Eval scores for the corresponding agent--criterion pair.}
\label{tab:qualitative_results_vertical}
\end{table}

Bland--Altman analysis in Figure~\ref{fig:bland-altman} further characterizes these differences by revealing a systematic \textit{positive bias} of G-Eval relative to humans, as well as an increasing dispersion from \textit{Outer Speech} to \textit{Scope Detection}.
This trend indicates that automatic judging can be informative for rapid regression testing of robot-to-human explanations.
However, it may be less reliable to assess compact control-oriented traces, that is, cases in which  perceived usefulness strongly depends on how well the fragment supports human interpretation of internal decisions under the given context.
In the Figure, we can observe that the gap is smallest and most stable for \textit{Outer Speech}.
This suggests that automatic judging is reasonably calibrated when evaluating explanations conditioned on explicit evidence.
In contrast, dispersion and bias increase markedly for \textit{Inner Speech} and peak for \textit{Scope Detection}, which indicates that judge models are less reliable on terse decision/diagnostic text, for which human judgments depend strongly on pragmatic usefulness and contextual fit rather than fluency.

As a complementary information, we compute BERTScore against curated reference sentences for each agent output~\cite{huang2026glenbenchgraphlanguagebasedbenchmark, zhang2025ngqa}.
\textit{Outer Speech} achieves the highest semantic similarity, with a score of $0.87$ (SD = $0.03$), which indicates a close alignment with the reference explanations.
\textit{Inner Speech} and \textit{Scope Detection} yield lower similarity scores, that is, $0.77$ (SD = $0.04$) and $0.77$ (SD = $0.03$), respectively, which is consistent with their more schematic and control-oriented role.
Since BERTScore measures semantic overlap rather than perceived interpretability, we report it as a separate alignment measure alongside the Likert-based qualitative evaluation.

\subsection{Discussion}
\label{sec:discussion}

The results provide two main takeaways about \textsf{JANUS}.
First, the \emph{factored controller} decomposition introduced in Section~\ref{sec:agent_model} yields a reliable \emph{control loop} in which intermediate variables, that is, domain selection, structured intent, readiness and tool-gating, evidence bundles, can be validated and measured in isolation.
Second, under the tested conditions, end-to-end responsiveness is dominated by LLM-based inference and prompt size, especially for format-critical stages, rather than by any single architectural bottleneck.

\paragraph{Reliability of the factored control loop}
Across the unit-test suites, all modules closely match curated references, as shown in Table~\ref{tab:module-metrics}.
This suggests that the typed interfaces and explicit gates introduced in Section~\ref{sec:agent_model} are effective at stabilizing long-horizon behavior.
\textit{Scope Detection} remains near-ceiling even when candidate domains have non-trivial semantic overlap, as shown in Fig.~\ref{fig:domain_similarity}, which indicates robustness to \emph{moderate} domain entanglement at the present scale.
Importantly, the $3$-domain similarity study does not show a monotonic degradation with overlap.
For the tested descriptions and model, errors appear to depend more on specific boundary cases than on average inter-domain similarity.
A stronger similarity effect is nonetheless plausible as the candidate set grows or domain definitions become more fine-grained, cases in which tighter decision boundaries and higher ambiguity would likely amplify confusions.

Once the active domain is fixed, \textit{Intent Recognition} and the execution-readiness role of \textit{Inner Speech} behave as intended.
Free-form human language is mapped to a structured task schema with typed arguments, and clarification is triggered when required slots are missing rather than being silently added by default.
\textit{Query Generation} produces executable grounding actions with high result agreement.
This supports the core architectural claim that \textsf{JANUS} can make \emph{groundedness} an enforceable property of the pipeline by routing answers through \textit{explicit} evidence acquisition and by conditioning \textit{Outer Speech} on the evidence bundle $B_t = (W_t, E_t)$, as provisioned in Equation~\eqref{eq:faithfulness}.

\paragraph{Latency as a property of the factorization}
The computational profiles in Figure~\ref{fig:boxplot} highlight that latency and token usage are not uniform across agents.
\textit{Query Generation} is the dominant cost driver, and exhibits the largest variance, which is expected given its long, format-constrained outputs and occasional multi-step retrieval plans.
\textit{Inner Speech}, while comparatively light in token usage, introduces a non-negligible compute cost because it performs deliberative validation over structured state and determines whether grounding can proceed at all.

A useful implication of factorization is that turn-level latency is \emph{scenario-dependent}, that is, it depends on which parts of the computational graph are actually activated.
Domain switching, clarification turns, and fully grounded answers correspond to distinct paths through Equations~\eqref{eq:scope}--\eqref{eq:outerspeech}, yielding qualitatively different response-time regimes.
This matters for HRI scenarios.
Even when median system-side delays are acceptable for assistant-style human-robot interactions, natural turn-taking is more timing-sensitive, and perceived responsiveness becomes an interaction-design variable rather than a pure compute optimization problem.
In practice, robot contingent behaviors, such as verbal fillers, gaze, or gestures, can partially decouple perceived responsiveness from raw compute time~\cite{backchanneling1, backchanneling2}.

\paragraph{Prompt length dominates cost in format-critical modules}
Token accounting indicates that most modules are strongly dominated by the input size.
Long instructions and exemplars are used to enforce strict structure and faithful communication, while the generated artifacts are comparatively short.
This is a double-edged property.
On the one hand, it is beneficial during development, because adding a small number of targeted exemplars can significantly stabilize structured outputs.
On the other hand, it also implies that prompt growth directly increases cost and latency.
In fact, as the number of domains and task tools scales, prompt management becomes central.
Practical mitigations include hierarchical routing, candidate-set pruning, constrained decoding, or task-specific fine-tuning for the most format-critical stages.

\paragraph{Separation of control traces and communication is empirically supported}
The qualitative study supports a key design principle of \textsf{JANUS}, that is, internal text channels as \textit{Scope Detection} and \textit{Inner Speech}, and the robot-to-human channel, that is, \textit{Outer Speech}, serve fundamentally different functions and should be optimized accordingly.
\textit{Outer Speech} is consistently perceived as clearer and more helpful, whereas control-oriented traces are less readable even when they are correct, because they encode compact state, slot diagnostics, and routing rationales rather than human-targeted explanations.
This suggests that exposing intermediate traces \textit{verbatim} should be treated as an explicit disclosure policy.
When internal traces are brought to surface, they likely require a \textit{translation layer} to preserve faithfulness while improving interpretability.

\paragraph{Automatic judging is useful but systematically optimistic}
Across agents, G-Eval tends to overestimate human-perceived quality, and the discrepancy increases as outputs become more control-oriented.
Bland--Altman analysis corroborates this pattern.
Agreement is most stable for robot-to-human explanations and degrades for terse decision/diagnostic text.
Practically, this suggests the use of automatic judging primarily for rapid iteration and regression testing on the \emph{communication} layer, while treating it as substantially less reliable for evaluating intermediate control traces, where human judgments depend on pragmatic usefulness and contextual fit beyond surface plausibility.

\paragraph{Semantic similarity is not explanation quality}
BERTScore provides content-alignment information against curated references, and is most informative for \textit{Outer Speech}, for which matching the intended answer and justification is a direct objective.
However, semantic similarity does not measure interpretability or faithfulness in the interaction sense.
A fragment can be \textit{semantically close} to a reference while still being judged only \textit{moderately helpful}, and control-oriented traces can be \textit{operationally useful} even when \textit{stylistically distant} from a human-authored reference.
For this reason, we treat BERTScore as a monitoring \textit{signal} for content drift, complementary to the Likert-based assessment of clarity, contextual relevance, and behavioral interpretability.

\paragraph{Limitations and next evaluation step}
The present evaluation uses curated unit-test suites that test each module in isolation, which is appropriate for validating the internal \textit{contracts} induced by the factorization but does not yet capture multi-turn feedback loops, recovery from upstream errors, or the dynamics of memory over long horizons.
The qualitative study is also limited in scale and is conducted outside an \textit{embodied} setting.
Perceived explanation quality in HRI depends on timing, prosody, embodiment cues, and disclosure policies.
A natural next step is therefore an end-to-end, multi-turn evaluation in an embodied deployment, in which humans experience grounding, clarification, and memory continuity across domain switches, and trust calibration can be assessed under realistic interaction dynamics.

\section{Conclusions}
\label{sec:conclusions}

This paper introduces \textsf{JANUS}, a cognitive architecture for assistive robots designed to support long-horizon human--robot interaction processes under partial observability.
As discussed in Section~\ref{sec:state_of_the_art}, current LLM-based robot cognitive architectures exhibit strong reasoning and tool-use capabilities, yet a persistent \textit{divide} remains between \emph{expressiveness} and \emph{control}.
Deliberation, tool invocation, and robot-to-human communication often collapse into a single generation step, which makes intermediate decisions difficult to verify before errors propagate.
\textsf{JANUS} addresses this gap by modeling the interaction process as a \emph{factored controller} with typed interfaces and explicit control variables.
In this case, routing, readiness, memory access, and evidence grounding can be inspected, audited, and measured.

The interaction process is formally modeled as a POMDP which latent state captures control-relevant variables such as domain, goal, task schema, parameter assignment, and persistent user context.
We formalized it as a factorization of the input--output mapping into a computational graph of modules, that is, \textit{Scope Detection}, \textit{Intent Recognition}, \textit{Memory}, \textit{Inner Speech}, \textit{Query Generation}, and \textit{Outer Speech}, together with explicit gates for sufficiency, readiness, and tool grounding.
Two architectural constructs are central to \textsf{JANUS}.
\emph{Inner Speech} acts as an internal control channel, since it validates the structured task representation, enforces parameter completeness, and decides whether the system should proceed, clarify, switch domain, or reject.
\emph{Memory} is implemented as a dedicated, stateful agent with a three-layer state and an explicit information-sufficiency gate to determine whether a bounded working context can be synthesized locally or must be augmented via archival retrieval.
Furthermore, \textsf{JANUS} supports runtime multi-domain behavior through a \textit{Customization Layer} that encapsulates domain descriptions, typed Task Tools, query-generation procedures, and plugins, without modifying core reasoning modules.

We evaluated \textsf{JANUS} using module-level unit tests on curated datasets in a safety-relevant dietary-assistance domain.
Across suites, the modules closely matched the curated references.
Latency profiles indicate that turn-level response time depends on the activated path through the factored controller.
The reported metrics are obtained from isolated unit tests, which are appropriate for validating the internal contracts induced by the factorization, but do not yet capture end-to-end error propagation, recovery from upstream mistakes, or long-horizon dynamics over multi-turn interaction.
Moreover, the evaluation focuses on a single domain and does not include longitudinal studies with real users or embodied deployment on physical robots, where timing, perception, and disclosure policies can materially affect trust and usability.

A future priority will be the end-to-end, longitudinal evaluation with humans to study how sufficiency gating, consolidation/revision policies, and \textit{Inner Speech} control decisions affect trust calibration, reliance, and the evolution of human models over time.

\section*{Acknowledgements}
This research is partially supported by the Italian government under the National Recovery and Resilience Plan (NRRP), Mission 4, Component 2, Investment 1.5, funded by the European Union NextGenerationEU programme, and awarded by the Italian Ministry of University and Research, project RAISE, grant agreement no. J33C22001220001. In addition, the ADVISOR project was founded by European Union - NextGenerationEU, Mission 4 Component 1 CUP E53D23016260001.

\bibliographystyle{elsarticle-num}
\bibliography{references}

\end{document}